%% file: main.tex
\documentclass[acmtog,review=false]{acmart}
\usepackage{booktabs} %

\usepackage{stackrel}
\usepackage{wasysym}
\usepackage{booktabs}
\newcommand{\new}[1]{\textcolor{black}{{#1}}}

\definecolor{tbcolor}{RGB}{10,110,210}

\newcommand{\etal}[0]{et al.}
\newcommand{\et}[0]{et}
\newcommand{\al}[0]{al.}

\usepackage[ruled]{algorithm2e} %

\SetAlFnt{\small}
\SetAlCapFnt{\small}
\SetAlCapNameFnt{\small}
\SetAlCapHSkip{0pt}

\acmJournal{TOG}
\acmVolume{39}
\acmNumber{6}
\acmArticle{223}
\acmMonth{12}

\setcopyright{rightsretained} 
\copyrightyear{2020} 
\acmYear{2020} 
\acmDOI{10.1145/3414685.3417803}

\begin{document}
\title{PIE: Portrait Image Embedding for Semantic Control}

\author{Ayush Tewari}
\email{atewari@mpi-inf.mpg.de}
\author{Mohamed Elgharib}
\email{elgharib@mpi-inf.mpg.de}
\author{Mallikarjun B R}
\email{mbr@mpi-inf.mpg.de}
\affiliation{%
 \institution{Max Planck Institute for Informatics, SIC}}
\author{Florian Bernard}
\email{fbernard@mpi-inf.mpg.de}
\affiliation{%
 \institution{Max Planck Institute for Informatics, SIC}}
\affiliation{%
 \institution{Technical University of Munich}}
\author{Hans-Peter Seidel}
\email{hpseidel@mpi-sb.mpg.de}
\affiliation{%
 \institution{Max Planck Institute for Informatics, SIC}
}
\author{Patrick P\'{e}rez}
\email{patrick.perez@valeo.com}
\affiliation{%
 \institution{Valeo.ai}}
\author{Michael Zollh\"{o}fer}
\email{zollhoefer@fb.com}
\affiliation{%
 \institution{Stanford University}}
\author{Christian Theobalt}
\email{theobalt@mpi-inf.mpg.de}
\affiliation{%
 \institution{Max Planck Institute for Informatics, SIC}}

\renewcommand\shortauthors{Tewari, A. et al}

\authorsaddresses{}




\begin{abstract}
\input{0_abstract.tex}
\end{abstract}

\begin{teaserfigure}
   \centering
   \includegraphics[width=0.9\textwidth]{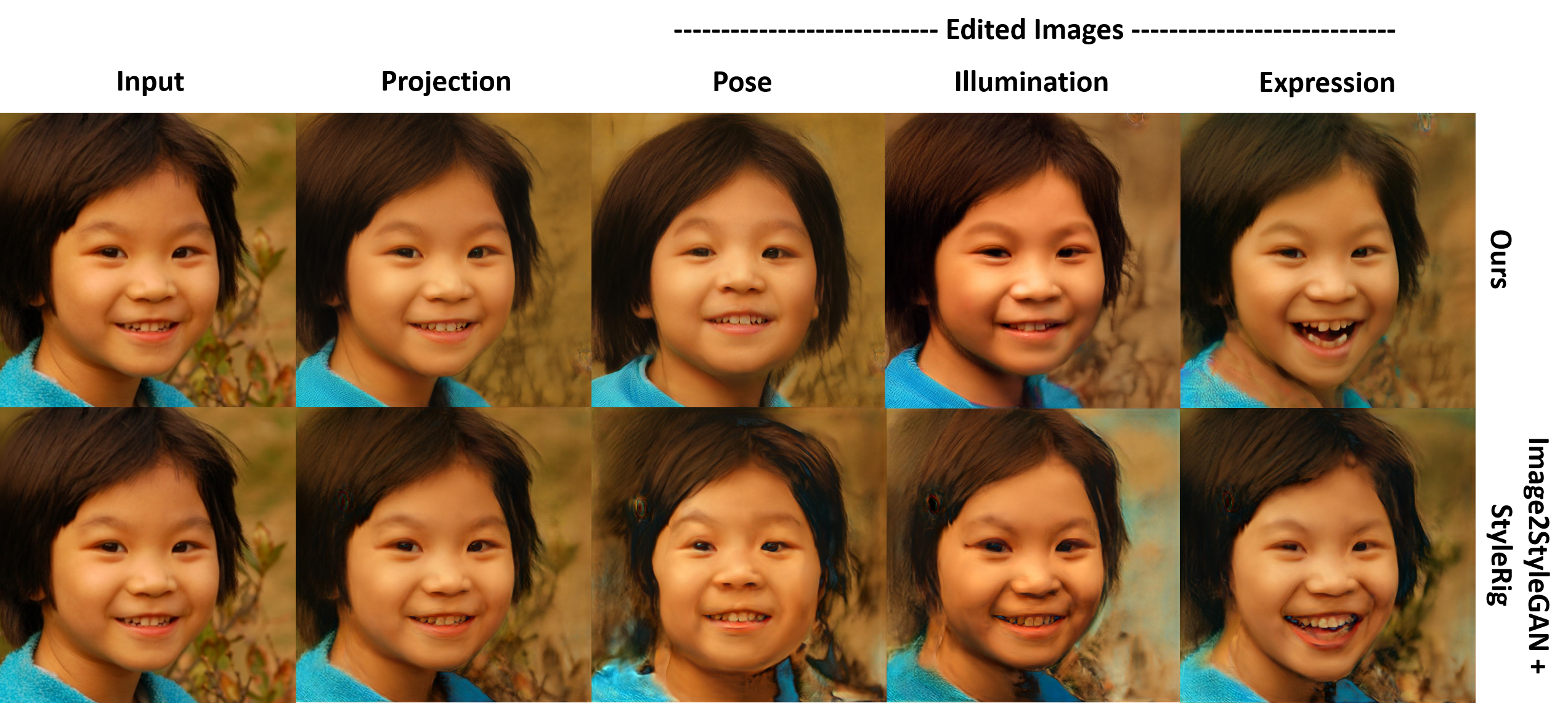}
   \caption{
   \new{We present an approach for embedding portrait images in the  latent space of StyleGAN~\cite{Karras2019cvpr} (visualized as ``Projection``) which allows for intuitive photo-real semantic editing of the head pose, facial expression, and scene illumination using StyleRig~\cite{anoynomous}.
   Our optimization-based approach allows us to achieve higher quality editing results compared to the existing embedding method Image2StyleGAN~\cite{Abdal_2019_ICCV}. Image from~\citet{shen2016deep}.      }
   }
   \label{fig:teaser}
\end{teaserfigure}

\maketitle

\input{1_intro_new_CT.tex}

\input{2_related}
\input{3_rigging}
\input{3_embedding}
\input{4_results}
\input{5_limitation}
\input{6_discussion}

\begin{acks}
We thank True-VisionSolutions Pty Ltd for providing the 2D face tracker.
We thank Rameen Abdal for kindly providing the \linebreak Image2StyleGAN code, Jalees Nehvi for helping us with the comparisons, and Gereon Fox for the video voiceover. 
This work was supported by the \grantsponsor{}{ERC}{} Consolidator Grant 4DReply (\grantnum{}{770784}), and the Max Planck Center for Visual Computing and Communications (MPC-VCC). 
We also acknowledge support from Technicolor and InterDigital.
\end{acks}

\bibliographystyle{ACM-Reference-Format}
\bibliography{main}

\end{document}

%% file: 0_abstract.tex
Editing of portrait images is a very popular and important research topic with a large variety of applications.
For ease of use, control should be provided via a semantically meaningful parameterization that is akin to computer animation controls.
\new{The vast majority of existing techniques do not provide such intuitive and fine-grained control, or only enable coarse editing of a single isolated control parameter.}
\new{Very recently, high-quality semantically controlled editing has been demonstrated, however only on synthetically created StyleGAN images.}
\new{We present the first approach for embedding real portrait images in the latent space of StyleGAN, which allows for intuitive editing of the head pose, facial expression, and scene illumination in the image.
Semantic editing in parameter space is achieved based on StyleRig, a pretrained neural network that maps the control space of a 3D morphable face model to the latent space of the GAN.}
\new{We design a novel hierarchical non-linear optimization problem to obtain the embedding.}
An identity preservation energy term allows spatially coherent edits while maintaining facial integrity.
Our approach runs at interactive frame rates and thus allows the user to explore the space of possible edits.
\new{We evaluate our approach on a wide set of portrait photos, compare it to the current state of the art, and validate the effectiveness of its components in an ablation study.}

%% file: 1_intro_new_CT.tex
\section{Introduction}
Portrait images, showing mainly the face and upper body of people, are among the most common and important photographic depictions.
We look at them to emotionally connect with friends and family, we use them to best present ourselves in job applications and on social media, they remind us of memorable events with friends, and photographs of faces are omnipresent in advertising.
Nowadays, tools to computationally edit and post-process photographs are widely available and heavily used.
Professional and hobby photographers use them to bring out the best of portrait and social media photos, as well as of professional imagery used in advertising.
Photos are often post-processed with the purpose to change the mood and lighting, to create a specific artistic look and feel, or to correct image defects or composition errors that only become apparent after the photo has been taken.
Today, commercial software\footnote{For example: \url{www.adobe.com/Photoshop}} or recent research software \cite{Luan2017DeepPS,Gatys} offers a variety of ways to edit the color or tonal characteristics of photos.
Some tools even enable the change of visual style of photos to match certain color schemes \cite{Luan2017DeepPS,Shih14}, or to match a desired painterly and non-photo-realistic style \cite{Gatys,selim16}.
In many cases, however, edits to a portrait are needed that require more complex and high-level modifications e.g. modifying head posture, smile or scene illumination after the capture.
Enabling such edits from a single photograph is an extremely challenging and underconstrained problem.
\new{This is because editing methods need to compute reliable estimates of 3D geometry of the person and  lighting in the scene.}
Moreover, they need to photo-realistically synthesize modified images of the person and background in a perspectively correct parallax-respecting manner, while inpainting disoccluding regions.

For ease of use, editing methods should use semantically meaningful parameterizations, which for the rest of the paper means the following: Head pose, face expression and scene lighting should be expressed as clearly disentangled and intuitive variables akin to computer animation controls, such as coordinates and angles, blendshape weights, or environment map parameterizations.
Existing methods to edit human portrait imagery at best achieve parts of these goals.
Some model-based methods to realistically edit human expression \cite{thies2019,thies2016face} and head pose \cite{kim2018DeepVideo} fundamentally require video input and do not work on single images.  
Other editing approaches are image-based and cannot be controlled by intuitive parametric controls~\cite{Geng2018WarpguidedGF,elor2017bringingPortraits,wang2018fewshotvid2vid,fewshot-neuraltalkingheads,Siarohin_2019_NeurIPS}, only enable editing of a single semantic parameter dimension, e.g., scene illumination \cite{Sun19,Meka:2019,Zhou_2019_ICCV}, or do not photo-realistically synthesize some important features such as hair~\cite{pagan}.

Recently, generative adversarial neural networks, such as StyleGAN~\cite{Karras2019cvpr}, were trained on community face image collections to learn a manifold of face images.
They can be sampled to generate impressive photo-realistic face portraits, even of people not existing in reality.
However, their learned parameterization entangles important face attributes 
(most notably identity, head pose, facial expression, and illumination), which thus cannot be independently and meaningfully controlled in the output. 
It therefore merely allows control on a coarse style-based level, e.g., to adapt or transfer face styles on certain frequency levels between images.
\new{To overcome this limitation, StyleRig~\cite{anoynomous} describes a neural network that maps the parameters of a 3D morphable face model (3DMM) \cite{Blanz1999} to a pretrained StyleGAN for face images.}
However, while their results show disentangled control of face images synthesized by a GAN, %
they do not allow for editing real portrait photos.

\new{On the other hand, some approaches have tried to embed real images in the StyleGAN latent space.
\citet{Abdal_2019_ICCV,abdal2019image2stylegan} demonstrate high-quality embedding results, which are used to perform edits such as style or expression transfer between two images, latent space interpolation for morphing, or image inpainting.
However, when these embeddings are used to edit the input images using StyleRig~\cite{anoynomous}, the visual quality is not preserved and the results often have artifacts.
High-quality parametric control of expression, pose or illumination on real images has not yet been shown to be feasible. }

\new{We therefore present the first method for embedding real portrait images in the StyleGAN latent space which allows for photo-realistic editing that combines all the following features:
It enables photo-real semantic editing of all these properties --- head pose, facial expression, and scene illumination, given only a single in-the-wild portrait photo as input, see Fig.~\ref{fig:teaser}.}
Edits are coherent in the entire scene and not limited to certain face areas.
Edits maintain perspectively correct parallax, photo-real occlusions and disocclusions, and illumination on the entire person, without warping artifacts in the unmodeled scene parts, such as hair.
The embedding is estimated based on a novel non-linear optimization problem formulation. 
Semantic editing in parameter space is then achieved based on the pretrained neural network of~\citet{anoynomous}, which maps the control space of a 3D morphable face model to the  latent space of StyleGAN.
These semantic edits are accessible through a simple user interface similar to established face animation control. 
We make the following contributions:
\begin{itemize}
    \item 
    \new{We propose a hierarchical optimization approach that embeds a portrait image in the latent space of StyleGAN while ensuring high-fidelity as well as editability.}
    \item 
    \new{Moreover, in addition to editability of the head pose, facial expression and scene illumination, we introduce an energy that enforces preservation of the facial identity. }
\end{itemize}

%% file: 2_related.tex
\section{Related Work}
We define face editing as the process of changing the head pose, facial expression, or incident illumination in a portrait image or video.
Many recent editing techniques are learning-based.
We distinguish between person-specific techniques that require a large corpus of images (or a long video) of the person, few-shot techniques that only require a small number of images, and single-shot techniques that only require a single image as input.
Our Portrait Image Embedding (PIE) approach is part of the third category and enables intuitive editing of a portrait image by a set of semantic control parameters. In addition to these categories, we will also summarize existing works related to portrait relighting.

\subsection{Person-specific Video Editing Techniques}
There has been a lot of research on person-specific techniques~\cite{thies2016face,kim2018DeepVideo,Recycle-GAN,thies2019,Kim19NeuralDubbing,
Wiles18} that require a large training corpus of the target person as input.
These approaches can be classified into model-based~\cite{thies2016face,kim2018DeepVideo,thies2019,Kim19NeuralDubbing} and image-based~\cite{Recycle-GAN} techniques.
Model-based techniques employ a parametric face model to represent the head pose, facial expression, and incident scene illumination.
The semantic parameter space spanned by the model can be used to either perform intuitive edits or transfer parameters from a source to a target video.
On the other end of the spectrum are image-based techniques that can transfer parameters, but do not provide intuitive semantic control.

\paragraph{Model-based Video Editing Techniques}
Facial reenactment approaches \cite{thies2016face,thies2019} change the facial expressions in a target video to the expressions in a driving source video.
These approaches achieve impressive results, but require a video of the target person as input and do not enable editing of the head pose and incident illumination.
Kim \etal~\shortcite{kim2018DeepVideo} proposed the first full head reenactment approach that is able to edit the head pose as well as the facial expression.
A conditional deep generative model is leveraged as a neural rendering engine.
While these approaches \cite{thies2016face,kim2018DeepVideo,thies2019} produce exciting results, they do not preserve the speaking style of the target.
In Kim~\etal~\shortcite{Kim19NeuralDubbing}, an approach is proposed for editing the expressions of a target subject while maintaining his/her speaking style.
This is made possible by a novel style translation network that learns a cycle-consistent mapping in blendshape space.
In contrast to our approach, all these techniques require a long video of the target as input and cannot edit a single image of an arbitrary person.

\paragraph{Image-based Video Editing Techniques}
Image-based techniques enable to control a target face through a driving video.
The approach of Bansal~\etal~\shortcite{Recycle-GAN} allows them to modify the target video while maintaining the speaking style.
A novel recycle loss is defined in the spatio-temporal video domain.
This approach obtains high-quality results for expressions and pose transfer.
In contrast to our approach, image-based approaches do not provide intuitive control via a set of semantic control parameters and have to be trained in a person-specific manner.
Thus, they cannot be employed to edit a single given image.

\subsection{Few-shot Editing Techniques}
Few-shot editing techniques \cite{wang2018fewshotvid2vid,fewshot-neuraltalkingheads,Wiles18} require only a small set of images of the target person as input.
Given multiple frames showing a target person, X2Face~\cite{Wiles18} drives a frontalized face embedding by a regressed warp field that is estimated by an encoder-decoder network.
The approach can also drive faces based on audio.
Wang~\etal~\shortcite{wang2018fewshotvid2vid} presented a few-shot video editing approach and showed its application to driving a target face via a source video.
A novel network weight generation module is proposed that is based on an attention mechanism.
To animate faces, the network is trained to transfer image sketches to photo-realistic face images.
The network is trained on a large multi-identity training corpus and can be applied to new unseen still images.
Zakharov~\etal~\shortcite{fewshot-neuraltalkingheads} presented a few-shot technique for animating faces.
Their solution has three components:
1) a generator network that translates landmark positions to photo-realistic images, 
2) an embedding network that learns an intermediate representation for conditioning the generator, and 
3) a discriminator.
The network is trained on a large corpus of face images across multiple identities and generalizes to new identities at test time.
Impressive results are shown in animating images, including legacy photos and even paintings.
The learned models of few-shot techniques \cite{fewshot-neuraltalkingheads,wang2018fewshotvid2vid,Wiles18} can be improved by fine-tuning on a few example images of the target person, e.g., images taken at different view-points or at different time instances.
The learned models can also be applied directly to new still images without fine-tuning.

\subsection{Single-shot Editing Techniques}
Several works~\cite{elor2017bringingPortraits,Geng2018WarpguidedGF,pagan} exist for controlling the expression and head pose given a single image as input.
Nagano~\et~\al~\shortcite{pagan} presented \textit{paGAN}, an approach for creating personalized avatars from just a single image of a person.
However, the work does not synthesize photo-realistic hair.
The approach of Averbuch-Elor~\etal~\shortcite{elor2017bringingPortraits} brings portrait images to life by animating their expression and pose.
The target image is animated through a 2D warp that is computed from the movement in the source video.
The mouth interior is copied from the source and blended into the warped target image.
The approach of Geng~\etal~\shortcite{Geng2018WarpguidedGF} 
employs deep generative models to synthesize more realistic facial detail and a higher quality mouth interior.
First, a dense spatial motion field is used to warp the target image.
Afterwards, the first network corrects the warped target image and synthesizes important skin detail.
Finally, the second network synthesizes the mouth interior, including realistic teeth.
Siarohin~\etal~\shortcite{Siarohin_2019_NeurIPS} proposed a method for animating a single image based on a driving sequence. By detecting keypoints in both the target image and the driving frames, the method uses a neural network to compute a dense warping field, specifying how to translate the driving frames into the target image. Based on this information a second network produces high-quality output frames. Since keypoint extraction is also learned during training, the method is applicable for any category of input, and in particular works for face  and full body images.
While existing single-shot editing techniques can only be controlled via a driving video, our approach enables intuitive editing of the head pose, facial expression and incident illumination in a portrait image 
through intuitive parametric control, as well as through a driving video.

\subsection{Portrait Relighting}
Relighting approaches modify the incident illumination on the face~\cite{Peers07,Shu17b,Zhou_2019_ICCV,Sun19,Meka:2019}. 
Earlier works~\cite{Peers07,Shu17b} require an exemplar portrait image that has been taken under the target illumination conditions.
More recent techniques use deep generative models~\cite{Zhou_2019_ICCV,Sun19,Meka:2019} and can relight images based on an environment map. 
Zhou~\et~\al~\shortcite{Zhou_2019_ICCV} train a relighting technique based on a large corpus of synthetic images.
Relighting is performed in the luminance channel, which simplifies the learning task.
Sun~\etal~\shortcite{Sun19} use light stage data to train their relighting approach.
At test time, the network produces high quality relighting results, even for in-the-wild images.
While training with light stage data leads to high-quality results, their scarcity and careful recording protocol can limit their adaptation.
Meka~\etal~\shortcite{Meka:2019} showed that the 4D reflectance field can be estimated from two color gradient images captured in a light stage.
This provides more movement flexibility for the subject during recording, and hence takes an important step towards capturing relightable video.

\new{ \subsection{Image Editing using SyleGAN}
Several recent methods have been proposed to edit StyleGAN generated images.}
\new{Most approaches linearly change the StyleGAN latent codes for editing~\cite{anoynomous,shen2020interpreting,hrknen2020ganspace}.
Non-linear editing has been shown in \citet{abdal2020styleflow}.
Image2StyleGAN~\cite{Abdal_2019_ICCV,abdal2019image2stylegan} is a popular approach for embedding real images into the StyleGAN latent space with very high fidelity. 
InterFaceGAN~\cite{shen2020interpreting} and StyleFlow~\cite{abdal2020styleflow} demonstrate editing of real images using these embeddings. 
Very recently, \citet{zhu2020indomain} introduce a domain-guided embedding method which allows for higher-quality editing, compared to Image2StyleGAN. 
However, they do not demonstrate results at the highest resolution for StyleGAN. 
In this paper, we design an embedding method which allows for high-quality portrait editing using StyleRig~\cite{anoynomous}.}

%% file: 3_rigging.tex
\section{Rigging StyleGAN-generated images}

StyleGAN~\cite{Karras2019cvpr} can synthesize human faces at an unprecedented level of photorealism. However, their edits are defined in terms of three main facial levels (coarse, medium and fine), with no semantic meaning attached to them. 
StyleRig~\cite{anoynomous} attaches a semantic control for a StyleGAN embedding, allowing edits for head pose, illumination and expressions. The control is defined through a 3D Morphable Face Model (3DMM)~\cite{Blanz1999}.

\subsection{StyleRig in more detail}
\label{sec:fmodel}
Faces are represented by a 3DMM model with $m = \text{257}$ parameters
\begin{equation}
\label{eq:parameters}
\theta = (\phi,\rho,\alpha,\delta,\beta,\gamma) \in \mathbb{R}^{257} \,.
\end{equation}
Here, $(\phi,\rho)\in \mathbb{R}^{6}$ are the rotation and translation parameters of the head pose, where rotation is defined using Euler angles. The vector $\alpha \in \mathbb{R}^{80}$ represents the geometry of the facial identity, while $\beta \in \mathbb{R}^{64}$ are the expression parameters. Skin reflectance is defined by $\delta \in \mathbb{R}^{80}$ and the scene illumination by $\gamma \in \mathbb{R}^{27}$. The basis vectors of the geometry and reflectance models are learned from 200  facial 3D scans \cite{Blanz1999}. The expression model is learned from FaceWarehouse \cite{Cao2014b} and the Digital Emily project \cite{alexander2010digital}. Principal Components Analysis (PCA) is used to compress the original over-complete blendshapes to a subspace of 64 parameters. Faces are assumed to be Lambertian, where illumination is modeled using second-order spherical harmonics (SH) \cite{Ramamoorthi2001b}.

StyleRig~\cite{anoynomous} allows one to semantically edit synthetic StyleGAN images.
To this end, StyleRig trains a neural network, called \emph{RigNet}, which can be understood as a function $\operatorname{rignet}(\cdot,\cdot)$ that maps a pair of StyleGAN code $\mathbf{v}$ and subset of 3DMM parameters $\theta^{\tau}$ to a new StyleGAN code $\hat{\mathbf{v}}$, i.e. $\hat{\mathbf{v}} = \operatorname{rignet}(\mathbf{v},\theta^{\tau})$. 
In practice, the 3DMM parameters are first transformed before being used in the network. Please refer to the supplemental document for details. 
With that, $\mathbf{I}_{\hat{\mathbf{v}}}$  shows the face of $\mathbf{I}_{\mathbf{v}}$ modified according to $\theta^{\tau}$ (i.e. with edited head pose, scene lighting, or facial expression), \new{where $\mathbf{I}_{\mathbf{v}}$ is the StyleGAN image generated using the latent code $\mathbf{v}$}. Thus, editing a synthetic image $\mathbf{I}_{\mathbf{v}}$ amounts to modifying the component $\tau$ in the parameter $\theta$, and then obtaining the edited image as $\mathbf{I}_{\hat{\mathbf{v}}} = \mathbf{I}(\operatorname{rignet}(\mathbf{v},\theta^{\tau}))$.
 Multiple RigNet models are trained, each to deal with just one mode of control (pose, expression, lighting). Although RigNet allows for editing of facial images, it has the major shortcoming that only \emph{synthetic} images can be manipulated, rather than real images. This is in  contrast to this work, where we are able to perform semantic editing of \emph{real} images.
\new{Different from the original RigNet design where a differentiable face reconstruction network regresses the 3DMM parameters from a StyleGAN code, we use a model-based face autoencoder~\cite{tewari17MoFA} which takes an image as an input. 
This change is necessary, as we initially do not have the StyleGAN code for the real image we want to edit.
}

%% file: 3_embedding.tex
\section{Semantic Editing of Real Facial Images}
We present an approach that allows for semantic editing of real facial images.
The key of our approach is to embed a given facial image in the StyleGAN latent space \cite{Karras2019cvpr}, where we pay particular attention to finding a latent encoding that is \emph{suitable for editing the image}. 
\new{This is crucial, since the parameter space of the StyleGAN architecture is generally under-constrained. For example, it has been shown that a StyleGAN trained for human faces is able to synthesize images that show completely different content with high fidelity, such as images of cat faces~\cite{Abdal_2019_ICCV}} %
\new{Our goal is to compute embeddings which can be edited using 3DMM parameters using StyleRig. }

\begin{figure*}
  \includegraphics[width=\textwidth]{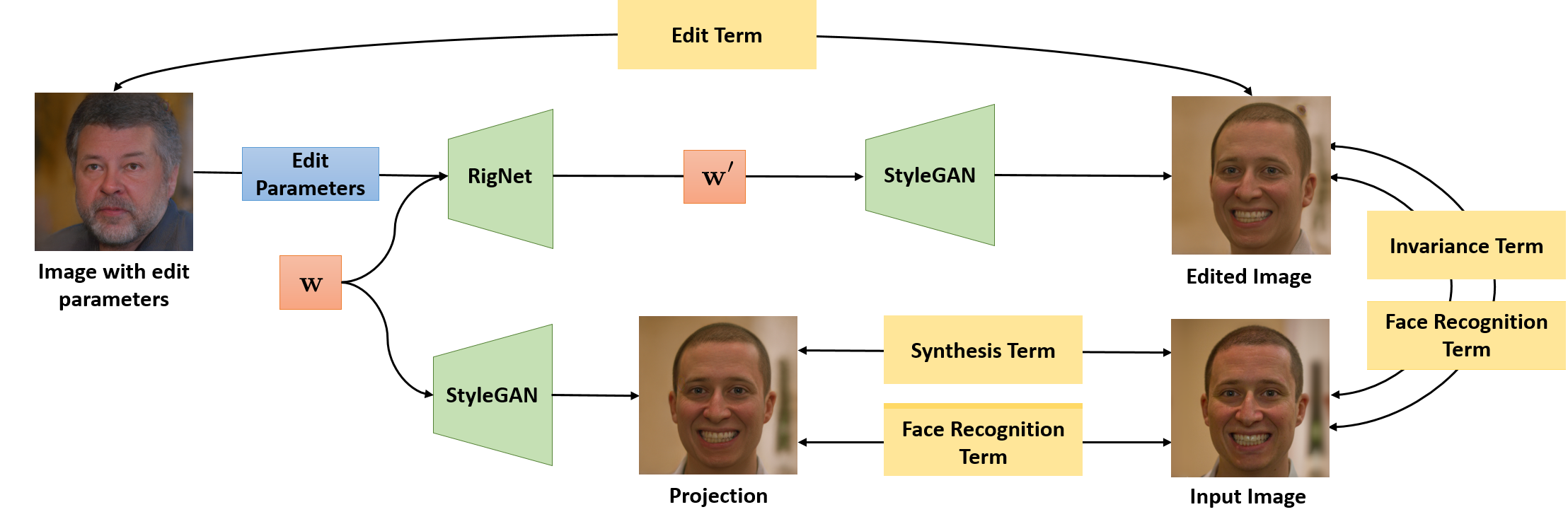}
  \caption{
Given a portrait input image, we optimize for a StyleGAN embedding which allows to faithfully reproduce the image (synthesis and facial recognition terms), editing the image based on semantic parameters such as head pose, expressions and scene illumination (edit and invariance terms), as well as preserving the facial identity during editing (facial recognition term).  
A novel hierarchical non-linear optimization strategy is used to compute the result. 
StyleGAN generated images (image with edit parameters) are used to extract the edit parameters during optimization. 
At ``test time'',~i.e. for performing portrait image editing, the image with edit parameters is not needed.
Note that the identity term is not visualized here. 
Images from~\citet{Shih14}.
}
  \label{fig:pipeline}
\end{figure*}

\paragraph{Problem Statement}
We will refer to the image that we want to make editable as $\mathbf{I}$ (without any subscripts or arguments), which we assume to be a given input.
Moreover, we will refer to the StyleGAN code that will make image $\mathbf{I}$ editable as $\mathbf{w}$, which is the desired output of our approach. As such, we will introduce an energy function $E(\mathbf{w})$, which is minimized by solving a numerical optimization problem. This energy function accounts for the high fidelity of the synthesized image based on $\mathbf{w}$ (explained in Sec.~\ref{sec:hfimsyn}), 
for editing-suitability (described in Sec.~\ref{sec:faceediting}), as well as for consistent face identity before and after the edit (Sec.~\ref{sec:recog}).
We emphasize that our approach is based on non-linear optimization techniques, and does not perform any learning of network weights, which in turn  means that we do not require any ground truth data of edited facial images.
In order to formulate the energy function we will make use of several existing neural networks, where all of them are pretrained and remain fixed throughout the optimization.
We will now introduce some technical notations, which will allow us to have an additional layer of abstraction and thereby facilitate a more comprehensive description of the main concepts.
\paragraph{Notation}
Throughout this paper we will use $\mathbf{w}$ exclusively to refer to the (unknown) StyleGAN embedding that we want to find, and we will use $\mathbf{v}$ (potentially with subscripts) to refer to general StyleGAN embeddings. We note that the  StyleGAN embeddings $\mathbf{w}$ and $\mathbf{v}$ can have two different forms, where each form has a different dimension, which we will describe in detail in Sec.~\ref{sec:opt}.
StyleGAN can be understood as a function $\operatorname{stylegan}(\cdot)$ that maps a given latent code  to a portrait image. To simplify notation, we will use the function notation $\mathbf{I}(\mathbf{v}) := \operatorname{stylegan}(\mathbf{v})$ in order to emphasize that we use the StyleGAN embedding $\mathbf{v}$ to generate the image $\mathbf{I}(\mathbf{v})$. Analogously, we overload $\mathbf{I}(\cdot)$, so that it can also take a 3DMM parameter $\theta$ as input. As such, $\mathbf{I}(\theta)$ refers to an image rendered using the face model that is parameterized by $\theta$ (Sec.~\ref{sec:fmodel}), where differentiable rendering is employed~\cite{tewari17MoFA}. Note that this rendered image is only defined on foreground face pixels as opposed to StyleGAN images.

 We use the variable $\tau \in \{\phi, \mathbb{\beta}, \gamma \}$ to indicate the user-defined facial semantic variable that is to be edited, which in our case can be the head pose $\phi$, facial expression $\beta$, or illumination $\gamma$. 
 Similarly, we use the complement notation~$\overline{\tau} \subset \{\phi, \rho, \alpha, \delta, \beta, \gamma\}$, to indicate all other variables, i.e., the ones that shall not be modified. With that, we use the notation $\theta^{\tau}$ (or $\theta^{\overline{\tau}}$) to refer to the extraction of the $\tau$-component (or $\overline{\tau}$-components) of $\theta$.
Since facial editing is implemented by modifying the $\tau$-component of the 3DMM parameter $\theta$, we write
$\theta' = [\theta_1^{\overline{\tau}}, \theta_2^{{\tau}}]$ to indicate that the respective $\tau$-component  of $\theta_1$ is replaced by the corresponding component in $\theta_2$. For example, for $\tau = \beta$,
\begin{align}
    \theta_1 &= (\phi_1,\rho_1,\alpha_1,\delta_1,\mathbb{\beta}_1,\gamma_1)\,,\quad\text{and}\\
     \theta_2 &= (\phi_2,\rho_2,\alpha_2,\delta_2,\mathbb{\beta}_2,\gamma_2)\,,\quad\text{we have}\\
     [\theta_1^{\overline{\tau}}, \theta_2^\tau] &= (\phi_1,\rho_1,\alpha_1,\delta_1,\mathbb{\beta}_2,\gamma_1) \,.
\end{align}

Moreover, we use the notation $\theta(\mathbf{v})$ to extract the 3DMM parameters from the StyleGAN embedding $\mathbf{v}$. In order to compute this, the embedding $\mathbf{v}$ is first used to synthesize the image $\mathbf{I}(\mathbf{v})$ (using StyleGAN),
followed by performing a 3D reconstruction based on the pretrained \emph{Model-based Face Autoencoder} (MoFA) network~\cite{tewari17MoFA}. Hence, for $\operatorname{MoFA}(\cdot)$ being the function that performs 3D reconstruction for a given image by estimating the 3DMM parameters, we define 
\begin{align}
    \theta(\mathbf{v}) = \operatorname{MoFA}(\mathbf{I}(\mathbf{v}))\,.
\end{align}
For any image $\mathbf{I'}$, we use the short-hand notation $\theta(\mathbf{I'}) = \operatorname{MoFA}(\mathbf{I'})$.
Similarly as above, we will use %
$\theta^{\tau}(\mathbf{v})$ and $\theta^{\tau}(\mathbf{I}')$ to extract only the $\tau$-component  from the 3DMM parameters. %
Whenever arguments of $\theta(\cdot)$ or $\mathbf{I}(\cdot)$ are fixed, i.e., the arguments are not a variable, we use the short-hand notations $\theta_{\mathbf{v}} = \theta(\mathbf{v})$, $\theta_{\mathbf{I}'} = \theta(\mathbf{I}')$, or $\mathbf{I}_{\mathbf{v}} = \mathbf{I}(\mathbf{v})$.
We summarize the most important part of our notations in Table~\ref{table:notation}.
\begin{table}[]
\caption{Summary of notation.}
\begin{tabular}{@{}ll@{}}
\toprule
 \textbf{Symbol} & \textbf{Meaning}  \\ \midrule
 $\mathbf{w}$ & StyleGAN embedding that we want to find  \\
 $\mathbf{v}$ & other StyleGAN embedding(s)  \\
  $\mathbf{\theta}$ & 3DMM parameter  \\
  $\mathbf{\tau}$ & component that is to be edited ($\tau \in \{\phi,\beta,\gamma\}$)  \\
 $\mathbf{I}$ & input image that we want to edit  \\
 $\mathbf{I}(\mathbf{v})$ &StyleGAN-synthesized image  \\
 $\mathbf{I}(\theta)$ & image of 3DMM rendering \\
  $\theta^{\tau} $ & extraction of $\tau$-component of $\theta$ \\ 
  $[\theta_1^{\overline{\tau}}, \theta_2^\tau]$ & combine   $\overline{\tau}$-components  in $\theta_1$  with  $\tau$-component in $\theta_2$ \\
  $\theta(\mathbf{v}), \theta_{\mathbf{v}}$ & 3D reconstruction of 3DMM parameters from  $\mathbf{I}(\mathbf{v})$ \\
  $\theta(\mathbf{I'}), \theta_{\mathbf{I'}}$ & 3D reconstruction  of 3DMM parameters from  $\mathbf{I}'$ \\
 \bottomrule
\end{tabular}
\label{table:notation}
\end{table}

\paragraph{Objective function}
We solve for $\mathbf{w}$ by minimizing the energy function
\begin{align}
\label{eq:totalenergy}
    E(\mathbf{w}) = E_{\text{syn}}(\mathbf{w}) +  E_{\text{id}}(\mathbf{w}) + E_{\text{edit}}(\mathbf{w}) +
    E_{\text{inv}}(\mathbf{w}) + E_{\text{recog}}(\mathbf{w}) \,.
\end{align}
$E_{\text{syn}}$ is a synthesis term enforcing the StyleGAN-synthesized image $\mathbf{I}(\mathbf{w})$ to be close to $\mathbf{I}$ (Sec.~\ref{sec:hfimsyn}). $E_{\text{id}}$, $E_{\text{edit}}$, $E_{\text{inv}}$ are face modification terms (Sec.~\ref{sec:faceediting}) enforcing edits to take place on the modified facial semantics while at the same time ensuring unmodified facial semantics to remain un-edited. $E_{\text{recog}}(\mathbf{w})$ is a face recognition term that will be introduced in Sec.~\ref{sec:recog}. A conceptual illustration of the energy function and the overall pipeline is shown in Fig.~\ref{fig:pipeline}. Next, we will discuss each term in more detail. 

\subsection{High-Fidelity Image Synthesis}
\label{sec:hfimsyn}
Similarly to Image2StyleGAN~\cite{Abdal_2019_ICCV}, we use the following energy term that accounts for the StyleGAN-synthesized image $\mathbf{I}(\mathbf{w})$  being close to $\mathbf{I}$:
\begin{align}
    E_{\text{syn}}(\mathbf{w}) & = \lambda_{\ell_2} \|{\mathbf{I}} - \mathbf{I}(\mathbf{w}) \|^2_2 + \lambda_{\text{p}} \| \mathbf{\Phi}({\mathbf{I}})-\mathbf{\Phi}( \mathbf{I}(\mathbf{w}) \|^2_2 \,.
    \label{eq:energy_synth}
\end{align}
The first term in the energy $E_{\text{syn}}$ penalizes the discrepancy between $\mathbf{I}$ and the synthesized image in terms of the (squared) $\ell_2$-norm, whereas the second term penalizes discrepancies based on the \emph{perceptual loss}~\cite{Johnson2016Perceptual}. 
The perceptual loss is estimated \new{on images downsampled by a factor of 4}, based on $\ell_2$-losses over VGG-16 layers \texttt{conv1\_1}, \texttt{conv1\_2}, \texttt{conv3\_2} and \texttt{conv4\_2} \cite{Simonyan15}. The notation $\mathbf{\Phi}(\cdot)$ refers to the function that downsamples a given input image and extracts features. The scalars $\lambda_{\ell_2}$ and  $\lambda_{\text{p}}$ are the relative weights of both terms.

In principle, we could minimize the energy $E_{\text{syn}}$ in~\eqref{eq:energy_synth} in order to obtain the StyleGAN code $\mathbf{w}$, as done in~\citet{Abdal_2019_ICCV}, and perform editing operations on $\mathbf{w}$. A so-obtained code vector $\mathbf{w}$ allows the use of StyleGAN to obtain a highly accurate synthetic version of the input face, which is even capable of reconstructing backgrounds with high accuracy. However, such a $\mathbf{w}$ is sub-optimal for performing \emph{semantic face editing}, as we later demonstrate in Fig.~\ref{fig:ablative-loss}. 
\subsection{Face Image Editing} \label{sec:faceediting}

\new{We augment the synthesis term with an editing energy that is based on the StyleRig framework~\cite{anoynomous}, which allows us to obtain more accurate semantic editing while preserving the non-edited attributes.}
Here, the StyleGAN embedding $\mathbf{w}$ that is to be determined should have the following three properties in order to be suitable for semantic editing:

\paragraph{Identity Property}
The identity property is phrased in terms of the $\ell_2$-norm of the difference of StyleGAN embeddings and is given by
\begin{align}
    E_{\text{id}}(\mathbf{w}) & =  \lambda_{\text{id}}\| \mathbf{w} - \operatorname{rignet}(\mathbf{w}, \theta^\tau(\mathbf{w})) \|_2^2 \,.
\end{align}
As such, whenever the RigNet is used to modify $\mathbf{w}$ with $\theta^\tau(\mathbf{w})$, i.e., a component of the 3DMM parameter extracted from $\mathbf{w}$, the embedding $\mathbf{w}$ should not be modified.

\paragraph{Edit Property}
In order to get around the obstacle of defining a suitable metric for 3DMM parameter vectors, whose components may be of significantly different scale, and the relative relevance of the individual components is not easily determined, we phrase the edit property in image space, \new{as in StyleRig~\cite{anoynomous}}.
As such, a facial edit is implicitly specified in image space via the StyleGAN embedding $\mathbf{v}$,
where
the $\tau$-component  of the respective 3DMM parameters of  $\mathbf{v}$, i.e. $\theta^{\tau}_{\mathbf{v}}$, specifies the  edit operation.
The image-space version of the edit property  reads
\begin{align}
\forall ~\mathbf{v}:\quad
    \mathbf{I}_{\mathbf{v}} =  \mathbf{I}(  [\theta_{\mathbf{v}}^{\overline{\tau}},
    \theta^{\tau}(\operatorname{rignet}(\mathbf{w}, \theta_{\mathbf{v}}^\tau)) ]) \,.
\end{align}
Note that this true equality cannot hold in practice, since the two images are from different domains (real image and a mesh rendering). 
We are interested in minimimzing the difference between these terms. 
This equation is best fulfilled whenever the $\tau$-component of the edited 3DMM parameters $\theta^\tau(\operatorname{rignet}(\mathbf{w}, \theta_{\mathbf{v}}^\tau))$ is equal to  $\theta_{\mathbf{v}}^\tau$, i.e. the edit has been successfully applied.
 Since computationally we cannot evaluate all choices of $\mathbf{v}$, we sample StyleGAN embeddings $\mathbf{v}$ as done in~\citet{anoynomous}, and then use the expected value as loss.
For integrating this property into our optimization framework we use a combination of a photometric term and a landmark term, which is defined as 
\begin{align}
    \ell(\mathbf{I}', \theta) = \lambda_{\text{ph}} \| \mathbf{I}' - \mathbf{I}(\theta)\|_{\text{\smiley{}}}^2 + \lambda_{\text{lm}} \| \mathcal{L}_{\mathbf{I}'} - \mathcal{L}(\theta)\|_F^2 \,.
\end{align}
The norm $\| \cdot \|_{\smiley{}}$ computes the $\ell_2$-norm of all \emph{foreground} pixels (the facial part of the image), whereas $\|\cdot\|_F$ is the Frobenius norm. By $\mathcal{L}_{\mathbf{I}'} \in \mathbb{R}^{66 \times 2}$ we denote the matrix of 2D facial landmarks extracted from the image $\mathbf{I}$ (based on~\citet{Saragih2011}), and $\mathcal{L}(\theta) \in \mathbb{R}^{66 \times 2}$ refers to the corresponding landmarks of the 3DMM after they have been projected onto the image plane.
With that, the edit property energy reads
\begin{align}
    E_{\text{edit}}(\mathbf{w}) & = 
    \lambda_{\text{e}} \, \mathbb{E}_{\mathbf{v}}[\ell( \mathbf{I}_{\mathbf{v}}, 
    [\theta_{\mathbf{v}}^{\overline{\tau}},
    \theta^{\tau}(\operatorname{rignet}(\mathbf{w}, \theta_{\mathbf{v}}^\tau))])] \,.
        \label{eq:edit}
\end{align}
    
\paragraph{Invariance Property}  
Similarly as the edit property we phrase the invariance property also in image space as%
\begin{align}
    \forall ~\mathbf{v}:\quad
    \mathbf{I} =  \mathbf{I}(
    [\theta^{\overline{\tau}}(\operatorname{rignet}(\mathbf{w}, \theta^\tau_{\mathbf{v}})) , \theta^{\tau}_{\mathbf{I}}]) \,.
\end{align}
While the edit property imposes that the $\tau$-component of the edited 3DMM parameter $\theta^\tau(\operatorname{rignet}(\mathbf{w}, \theta^\tau_{\mathbf{v}}))$ is modified as desired, the invariance property takes care of all  $\overline{\tau}$. It is fulfilled whenever it holds that $\theta^{\overline{\tau}}(\operatorname{rignet}(\mathbf{w}, \theta^\tau_{\mathbf{v}})) = \theta^{\overline{\tau}}_{\mathbf{I}}$, i.e. the components $\overline{\tau}$ that are not to be edited are maintained from the input image $\mathbf{I}$.

Analogously to the edit property, we base the respective energy on the combination of a photometric term and a landmark term as implemented by $\ell(\cdot)$, so that we obtain
\begin{align}
    E_{\text{inv}}(\mathbf{w}) & = \lambda_{\text{inv}} \, \mathbb{E}_{\mathbf{v}}[\ell(\mathbf{I}, [\theta^{\overline{\tau}}(\operatorname{rignet}(\mathbf{w}, \theta^\tau_{\mathbf{v}})), \theta^{\tau}_{\mathbf{I}}])] \,.
\end{align}
\subsection{Face Recognition Consistency}\label{sec:recog}
\new{In addition to the synthesis and editing terms, we incorporate two face recognition consistency terms to preserve the facial integrity while editing.}
 On the one hand, it is desirable that the synthesized image $\mathbf{I}(\mathbf{w})$ is recognized to depict the same person as shown in the given input image $\mathbf{I}$. 
 On the other hand, the edited image, $\operatorname{stylegan}(\operatorname{rignet}(\mathbf{w}, \theta^{\tau}_{\mathbf{v}}))$
should also depict the same person as shown in the input $\mathbf{I}$. 

In order to do so, we use VGG-Face~\cite{Parkhi15} to extract \emph{face recognition features},
where we use the notation $\mathbf{\Psi}(\cdot)$ to refer to the function that extracts such features from a given input image. We define the recognition loss
\begin{align}
\ell_{\text{recog}}(\mathbf{I}', \mathbf{v}) = \|\mathbf{\Psi}(\mathbf{I}') - \mathbf{\Psi}(\mathbf{I}(\mathbf{v})) \|_F^2 \,,
\end{align}
which is then used to phrase the recognition energy term as
\begin{align}
    \label{eq:recog}
    E_{\text{recog}}(\mathbf{w}) = \lambda_{\text{r}_{\mathbf{w}}} \,\ell_{\text{recog}}(\mathbf{I}, \mathbf{w}) 
    + \lambda_{\text{r}_{\hat{\mathbf{w}}}} \,\mathbb{E}_{\mathbf{v}}[\ell_{\text{recog}}(\mathbf{I}, \operatorname{rignet}(\mathbf{w}, \theta^{\tau}_{\mathbf{v}}))] \,.
\end{align}
\begin{figure*}
\includegraphics[width=\textwidth]{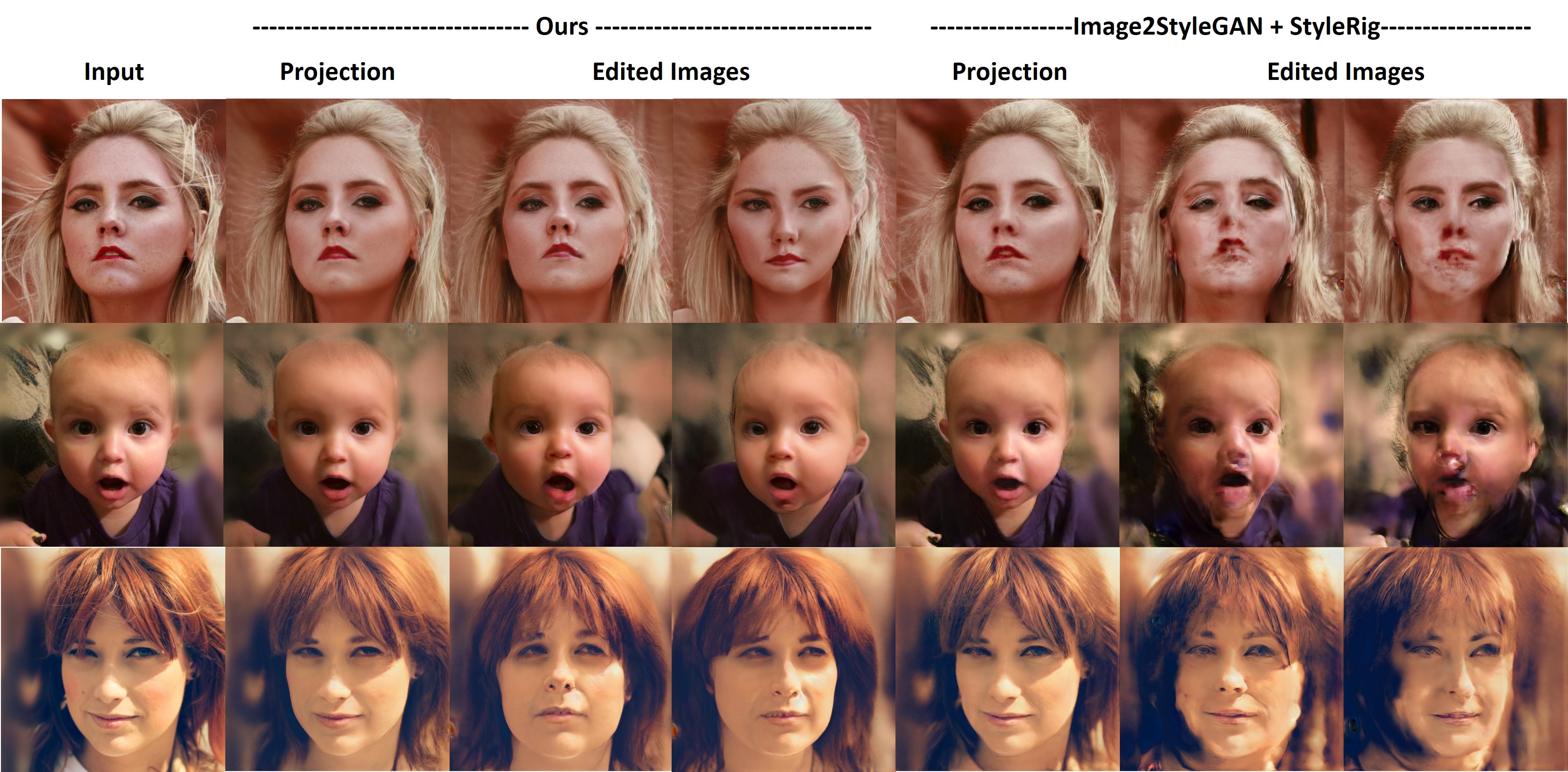}
\vspace{-0.6cm}
\caption{
\new{Pose Editing.
Our approach can handle a large variety of head pose modifications including out-of-plane rotations in a realistic manner.
Image2StyleGAN~\cite{Abdal_2019_ICCV} embeddings often lead to artifacts when edited using StyleRig. Images from~\citet{shen2016deep}.} %
}
\label{fig:pos-main}
\end{figure*}
\begin{figure*}
\includegraphics[width=\textwidth]{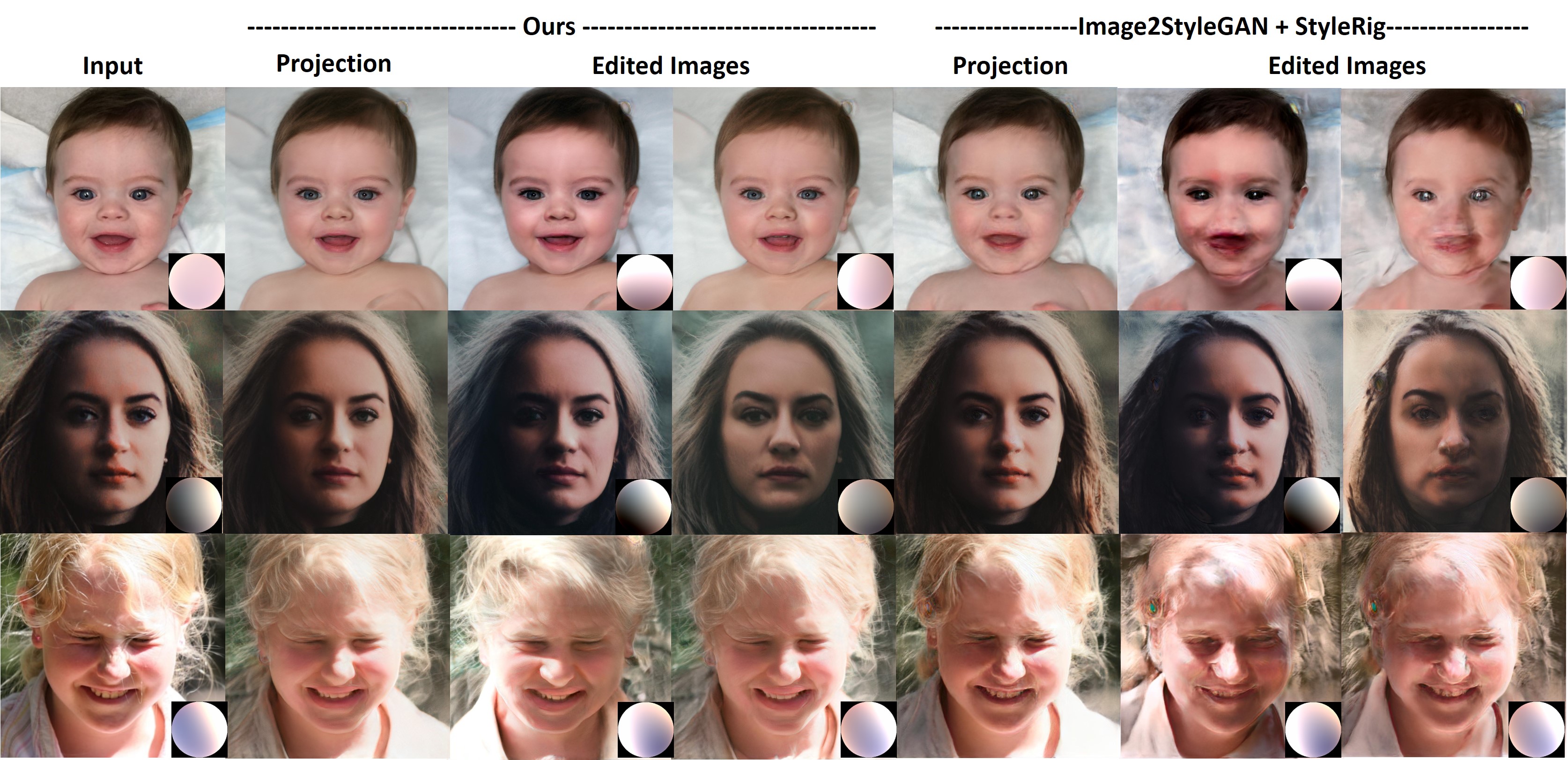}
\vspace{-0.6cm}
\caption{
\new{Illumination Editing.
Our approach can realistically relight portrait images.
Each edited image corresponds to changing a different Spherical Harmonics coefficient, while all other coefficients are kept fixed.
The environment maps are visualized in the inset.
Image2StyleGAN~\cite{Abdal_2019_ICCV} embeddings often lead to artifacts when edited using StyleRig. Images from~\citet{shen2016deep}.}
}
\label{fig:light-main-ne}
\end{figure*}
\begin{figure}
\includegraphics[width=0.48\textwidth]{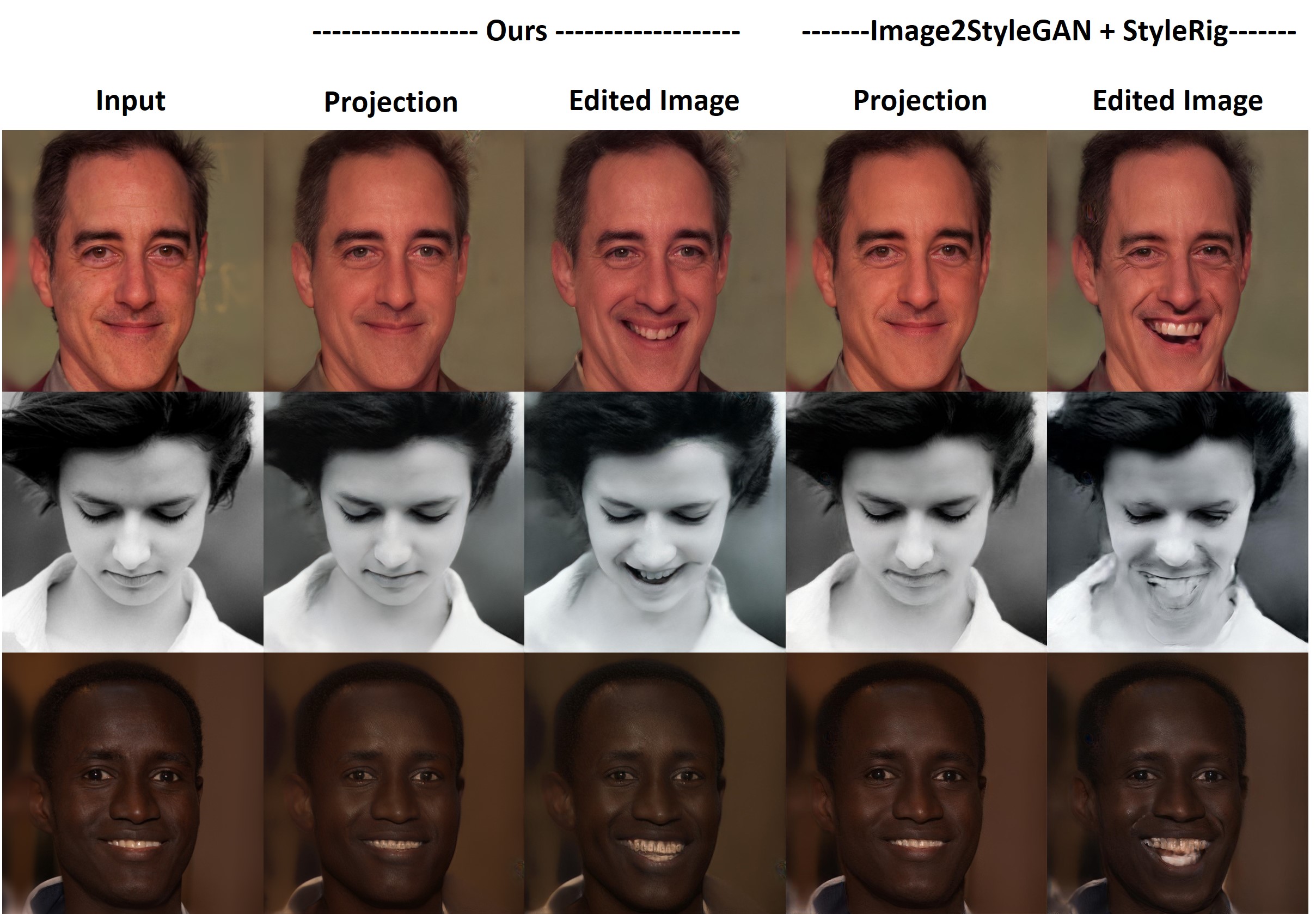}
\vspace{-0.6cm}
\caption{
\new{Expression Editing.
Our approach can also be used to edit the facial expressions in a portrait image in a realistic manner.
We obtain more plausible results, compared to Image2StyleGAN~\cite{Abdal_2019_ICCV} embeddings.  Images from~\citet{shen2016deep} and~\citet{Shih14}.}
}
\label{fig:exp-main}
\end{figure}

\subsection{Optimization}
\label{sec:opt}
Our energy function $E(\cdot)$ in \eqref{eq:totalenergy} depends on a range of highly non-linear functions, such as $\operatorname{stylegan}(\cdot)$, $\operatorname{MoFA}(\cdot)$, $\Phi(\cdot)$ and $\Psi(\cdot)$, which are implemented in terms of (pretrained) neural networks.
We implement our energy minimization within TensorFlow~\cite{tensorflow2015-whitepaper} using ADADELTA optimization~\cite{zeiler2012adadelta}. In each iteration we stochastically sample a different $\mathbf{v}$. 
The optimization uses a hierarchical approach that we describe next.

\paragraph{Hierarchical Optimization}
StyleGAN is based on a hierarchy of latent spaces, where a stage-one embedding $Z$ with $|Z|=512$ is randomly sampled first. This is then fed into 
a mapping network that produces $W$ as  output, where $|W|=512$. 
Subsequently, $W$ is  extended to $W^+$, where $|W^+| =18 \times 512$, and used as input to $18$ network layers. 
It has been shown that $W^+$ is the most expressive space for fitting to real  images~\cite{Abdal_2019_ICCV}.
However, we found that a direct optimization over this space leads to lower-quality editing results with severe artifacts. This is because the optimized variable can be far from the prior distribution of StyleGAN. To address this, we first optimize for the embedding in the $W$-space, meaning that in the first stage of our optimization the variable $\mathbf{w}$ is understood as an embedding in the $W$-space.
We optimize in $W$-space for $2000$ iterations. 
We then transfer the result to $W^+$-space,    initialize the variable $\mathbf{w}$ respectively, and
continue optimizing in the $W^+$-space for another $1000$ iterations. 
Optimizing in this hierarchical way allows us to represent the coarse version of the image in the $W$-space, which is less expressive and thereby closer to the prior distribution. 
Finetuning on the $W^+$ space then allows us to fit the fine-scale details, while preserving editing quality. 

%% file: 4_results.tex
\section{Results}
In the following, we demonstrate the high-quality results of our method, analyze its different components, as well as compare to several state-of-the-art approaches for portrait image editing.

\paragraph{Implementation Details}

We use the following weights for our energy terms: $\lambda_{\ell_2} = 10^{-6}$, $\lambda_{\text{p}} = 10^{-6}$,  $\lambda_{\text{id}} = 1.0$, $\lambda_{\text{ph}} = 0.001$, $\lambda_{\text{lm}} = 0.2$,  $\lambda_{\text{e}} = 10.0$, $\lambda_{\text{inv}} = 10.0$, $\lambda_{\text{r}_{\mathbf{w}}} = 0.1$, $\lambda_{\text{r}_{\hat{\mathbf{w}}}} = 0.1$.
We use a starting step size of $50$ when optimizing over embeddings in $W$ space, and $10$ in $W^+$ space. 
The step size is then exponentially decayed by a factor of $0.1$  every $2000$ steps.
Optimization takes approximately $10$ minutes for $3000$ iterations per image on an NVIDIA V100 GPU. 
Once the embedding is obtained, the portrait image can be edited at an interactive speed.
\new{\paragraph{Feedback}
\label{sec:feedback}
We noticed that a simple feedback loop allows us to get more accurate editing results. 
We update the parameters used as input to RigNet  in order to correct for the editing inaccuracies in the output. 
Given target 3DMM parameters $\mathbf{\theta}$, we first obtain the embedding for the edited image, $\operatorname{rignet}(\mathbf{w}, \theta^\tau)$. 
We then estimate the 3DMM parameters from the edited embedding, $\theta_{\text{est}} = \theta(\operatorname{rignet}(\mathbf{w}, \theta^{\tau}))$. %
The final embedding is computed as $\operatorname{rignet}(\mathbf{w}, \theta_{\text{new}}^\tau)$ with $\theta_{\text{new}} = \mathbf{\theta} + (\mathbf{\theta} - \mathbf{\theta_{\text{est}}})$.
}

\subsection{High-Fidelity Semantic Editing}
We evaluate our approach on a large variety of portrait images taken from~\citet{shen2016deep} and~\citet{Shih14}. 
The images are preprocessed as in StyleGAN~\cite{Karras2019cvpr}. 
Figs.~\ref{fig:pos-main},~\ref{fig:light-main-ne},~\ref{fig:exp-main}
show results of controlling the head pose,  scene illumination, and facial expressions, respectively. 
\new{The projections onto the StyleGAN space are detailed, preserving the facial identity.}
Our approach also produces photo-realistic edits.
Fig.~\ref{fig:pos-main} shows that our approach can handle a large variety of head pose modifications, including out-of-plane rotations. 
It also automatically inpaints uncovered background regions in a photo-realistic manner. 
Fig.~\ref{fig:light-main-ne} demonstrates our relighting results. Our approach can handle complex light material interactions, resulting in high photo-realism.
The relighting effects are not restricted to just the face region, with hair and even eyes being relit.
Our approach also allows for editing facial expressions, see Fig.~\ref{fig:exp-main}. 
For smooth temporal editing results of portrait images, please refer to the supplementary video. 
\begin{figure*}[t]
\includegraphics[width=\textwidth]{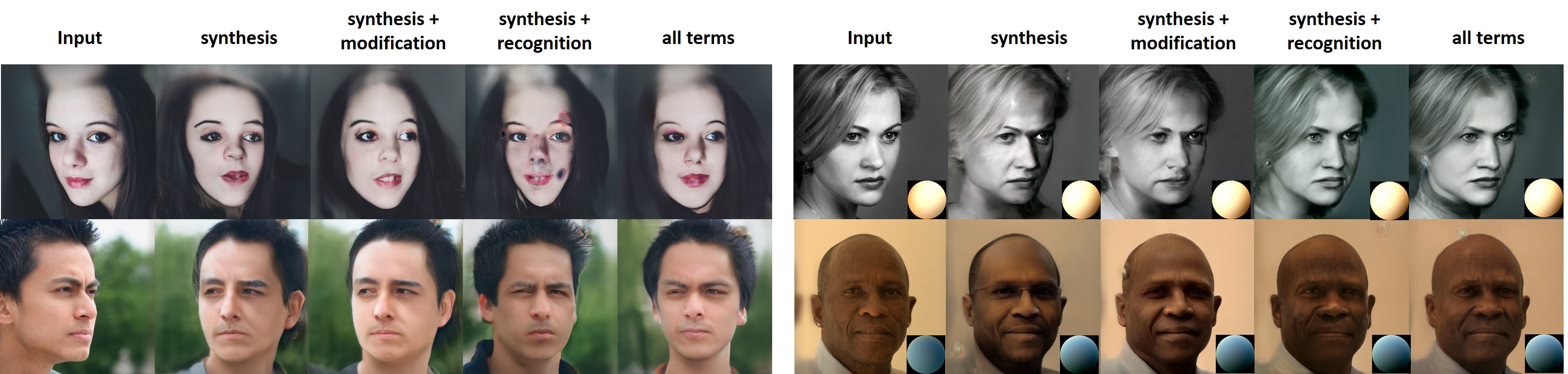}
\vspace{-0.6cm}
\caption{
Ablative analysis of the different loss functions.
\emph{Modification} refers to the edit, invariance and identity terms simultaneously.
The left block shows results for editing the head pose and the right block shows results for editing scene illumination. 
All losses are required to obtain high-fidelity edits.  Images from~\citet{shen2016deep}.
}
\label{fig:ablative-loss}
\end{figure*}

\begin{figure*}
\includegraphics[width=\textwidth]{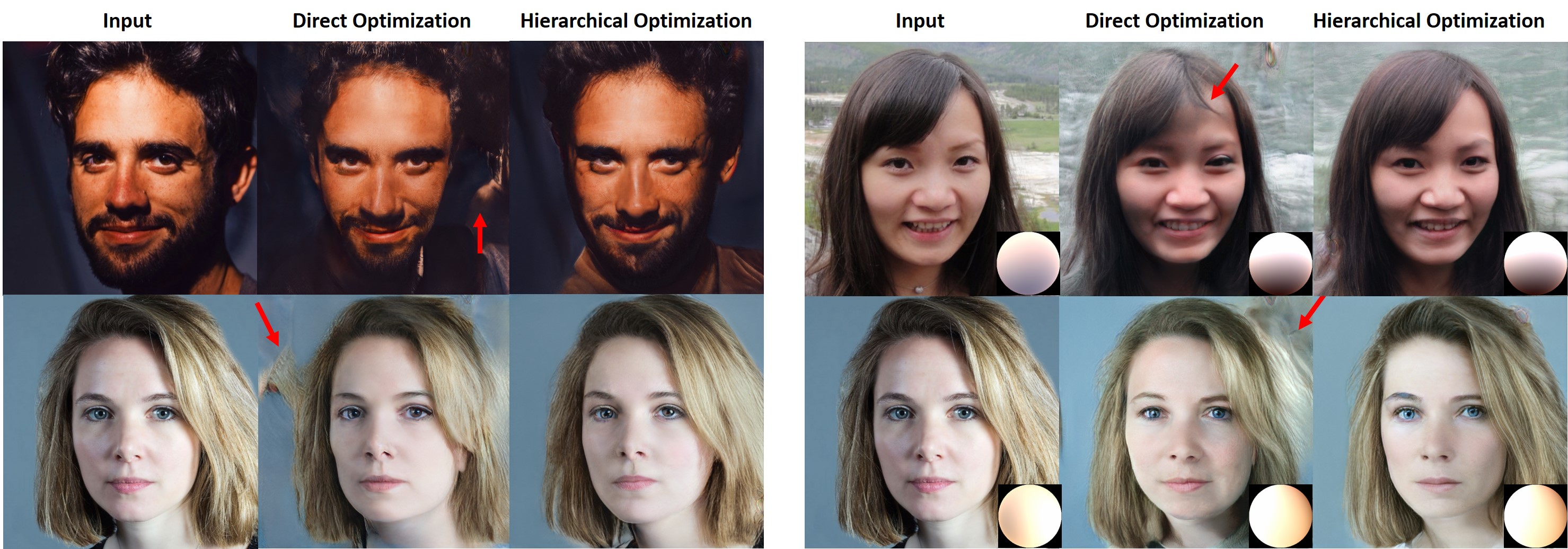}
\vspace{-0.6cm}
\caption{
Ablative analysis with and without hierarchical optimization. 
The left block shows the results for pose editing and the right block for illumination editing. 
Without our hierarchical optimization, the obtained embedding cannot be easily edited and artifacts appear in the modified images.  Images from~\citet{shen2016deep}.
} 
\label{fig:ablative-opt}
\end{figure*}

\begin{table*}[]
    \caption{ 
        We compare different settings quantitatively using several metrics for pose editing. 
        \new{All numbers are averaged over more than $2500$ pose editing results.
        We measure the quality of the fit by comparing them to the input image using PSNR and SSIM metrics.
        Editing error is measured as the angular difference between the desired and achieved face poses. 
        Recognition error measures the value of the facial recognition error for the edited images.
        There is usually a trade-off between the quality and accuracy of editing, as lower recognition errors correspond to higher editing errors.
        We also compare to Image2StyleGAN~\cite{Abdal_2019_ICCV} embeddings using these metrics. 
        While it achieves the highest quality fitting, the editing results do not preserve the facial identity well. }
        }
\begin{tabular}{@{}lcccccc}
\cmidrule(r){1-7}
 &
  \multicolumn{1}{c}{synthesis} &
      \begin{tabular}{@{}c@{}}synthesis + \\ recognition\end{tabular} &
  \begin{tabular}{@{}c@{}}synthesis + \\ modification\end{tabular} &
  \begin{tabular}[c]{@{}c@{}}all terms (PIE)\end{tabular}  &
  \begin{tabular}[c]{@{}c@{}}all terms (direct opt.)\end{tabular} &
  Image2StyleGAN \\ \cmidrule(r){1-7} 
PSNR (dB) $\uparrow$ / SSIM $\uparrow$   & 30.15 / 0.70 & 29.84 / 0.69 & 30.15 / 0.70 & 29.96 / 0.70 & 29.76 / 0.69 & \textbf{31.21} / \textbf{0.75} \\
Editing Error (rad) $\downarrow$ & 0.06         & 0.11         & \textbf{0.036}        & 0.08         & 0.037        & 0.07         \\
Recognition Error $\downarrow$   & 95.76        & 43.64        & 90.10        & \textbf{42.82}        & 51.65        & 275.40      \\ \bottomrule
\end{tabular}
\label{tab:ablative}
\end{table*}

\subsection{Ablation Studies}
Here, we evaluate the importance of the different proposed loss functions, and also evaluate the hierarchical optimization strategy. Please refer to the supplemental document for the evaluation of the feedback strategy. 

\paragraph{Loss Functions}
Fig.~\ref{fig:ablative-loss} shows  qualitative ablative analysis for the different loss functions. 
We group the edit, invariance and identity terms as \emph{modification terms}. 
Adding face recognition consistency without the modification terms lead to incorrect editing in some cases. 
Adding the modification terms without face recognition consistency leads to the method being able to accurately change the specified semantic property, but the identity of the person in the image is not preserved. 
Using all terms together leads to results with photorealistic edits with preservation of identity. 
We do not evaluate the importance of the individual components of the modification terms, as it was already evaluated in \citet{anoynomous}.

\paragraph{Hierarchical Optimization}
Hierarchical optimization is an important component of our approach.
Fig.~\ref{fig:ablative-opt} shows results with and without this component.
Without hierarchical optimization, the method directly optimizes for $\mathbf{w} \in W^+$.
While this leads to high-quality fits, the obtained embedding can be far from the training distribution of StyleRig.
Thus, the quality of edits is poor.
For example in Fig.~\ref{fig:ablative-opt} (top), the StyleGAN network interprets the ears as background, which leads to undesirable distortions.
With hierarchical optimization, the results do not suffer from artifacts. 
\paragraph{Quantitative Analysis} We also analyze the effect of different design choices quantitatively, see Tab.~\ref{tab:ablative}.
We look at three properties, the quality of recostruction (measured using \new{PSNR and SSIM between the projected image and the input)}, the accuracy of edits \new{(measured as the angular distance between the desired and estimated head poses)}, and idenity preservation under edits \new{(measured using the second term in Eq.~\ref{eq:recog})} during editing.
The numbers reported are averaged over more than \new{$2500$} pose editing results. 
We can see that removing the recognition term changes the identity of the face during editing, and removing the modification loss increases the editing and recognition error.
Hierarchical optimization also leads to better facial identity preservation, compared to direct optimization. 
This is expected, since the results with direct optimization often have artifacts. 
Note that the artifacts outside of the face region (hair, ears) would not increase the recognition errors significantly. 
\new{The recognition term introduces a clear trade-off between the quality of identity preservation under edits and the accuracy of edits.
The modification terms allow for slight improvements in both identity preservation as well as the accuracy of the edits. 
}

\subsection{Comparison to the State of the Art}
\begin{figure*}
\includegraphics[width=\textwidth]{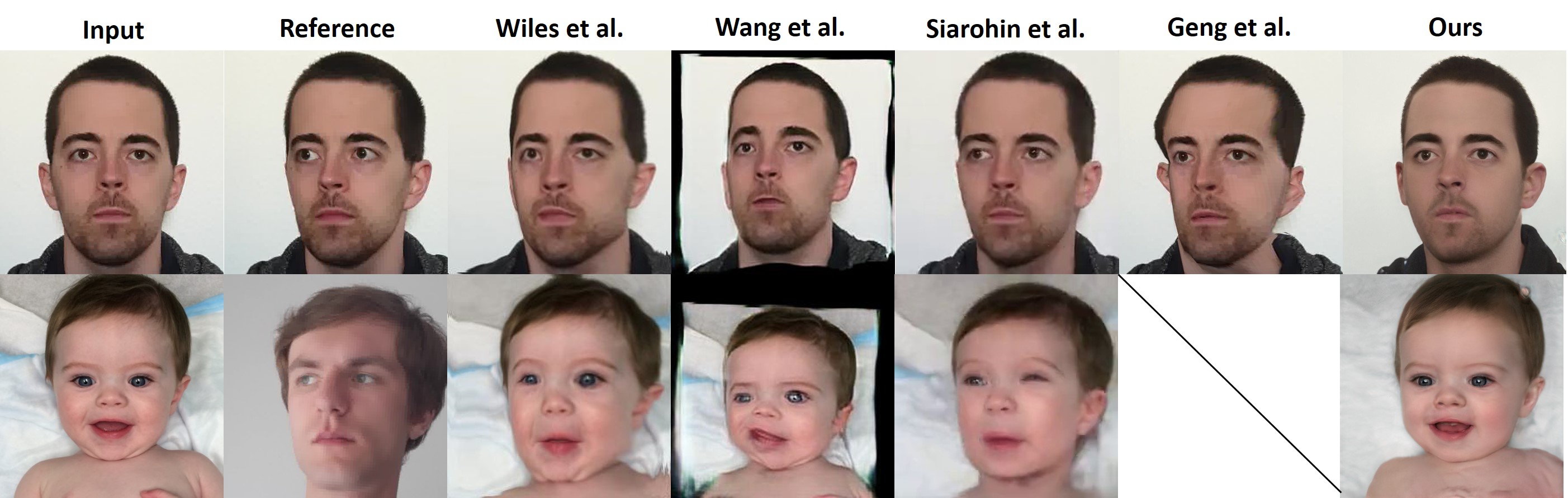}
\vspace{-0.3cm}
\caption{
Comparison of head pose editing for self-reenactment (first two rows) and cross-identity reenactment (last two rows).
We compare our approach to Wiles~\etal~\shortcite{Wiles18}, Wang~\etal~\shortcite{wang2018vid2vid}, Siarohin~\etal~\shortcite{Siarohin_2019_NeurIPS} and Geng~\etal~\shortcite{Geng2018WarpguidedGF}.
The pose from the reference images is transferred to the input. 
Our approach obtains higher quality head pose editing results, specially in the case of cross-identity transfer. 
All approaches other than ours are incapable of \emph{disentangled} edits, i.e., they cannot transfer the pose without also changing the expressions. 
The implementation of Geng~\etal~\shortcite{Geng2018WarpguidedGF} does not handle cross-identity reenactment. 
Note that while the three competing approaches require a reference image in order to generate the results, we allow for explicit control over the pose parameters.  Image from~\citet{shen2016deep}.
}
\label{fig:pose-comparison}
\end{figure*}
\begin{figure}
\includegraphics[width=0.49\textwidth]{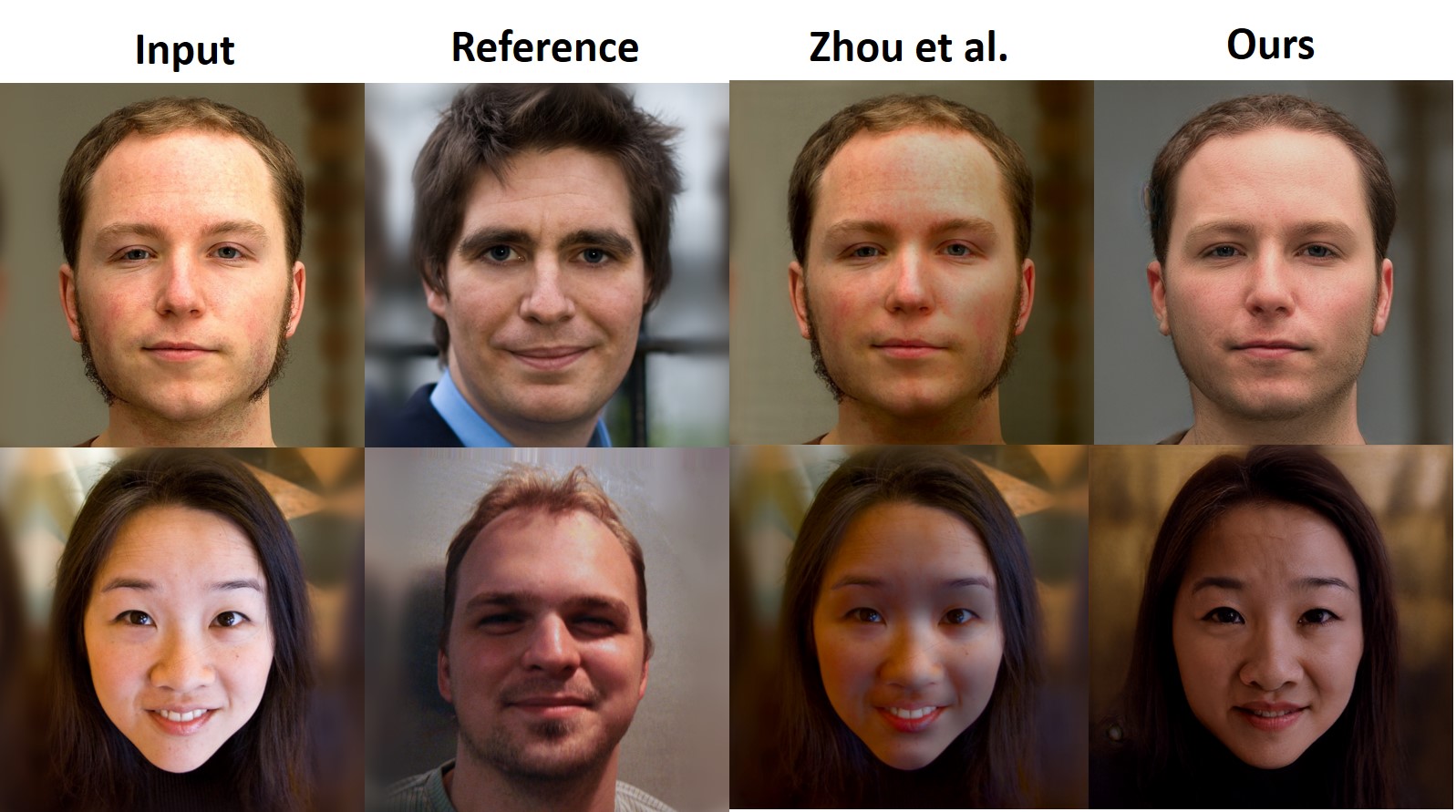}
\vspace{-0.6cm}
\caption{
Comparison of our relighting results with \citet{Zhou_2019_ICCV}. 
The illumination in the reference image is transferred to the input. 
Our results are more natural and achieve more accurate relighting. 
We can edit colored illumination while \citet{Zhou_2019_ICCV} can only edit monochrome light.
In addition, we can also edit the head pose and facial expressions, while~\citet{Zhou_2019_ICCV} is trained only for relighting. 
Images from~\citet{Shih14}.
}
\label{fig:light-comparison}
\end{figure}
\subsubsection{Image2StyleGAN}
\new{Image2StyleGAN~\cite{Abdal_2019_ICCV} also projects real images to the StyleGAN latent space, and is thus a closely related approach. 
The source code of Image2StyleGAN was kindly provided by the authors.
We show editing results using Image2StyleGAN embeddings in Figs.~\ref{fig:teaser},~\ref{fig:pos-main},~\ref{fig:light-main-ne} and ~\ref{fig:exp-main}. 
Since these embeddings are optimized only using the synthesis terms and without using hierarchical optimization, the results are often implausible, as is most evident when editing the head pose and scene illumination. 
However, Image2StyleGAN projections are more detailed than ours.}
\new{We also quantitatively compare to Image2StyleGAN in Tab.~\ref{tab:ablative}.
Image2StyleGAN obtains the highest quality projections in terms of PSNR and SSIM.
When combined with StyleRig, it also leads to low editing errors. 
However, the recognition errors are very high due to the artifacts in the results, as shown in the qualitative results. 
}
\subsubsection{Other Aproaches}
We \new{also} compare our approach to a number of related techniques,  X2Face~\cite{Wiles18}, \citet{Geng2018WarpguidedGF} and \citet{Siarohin_2019_NeurIPS}. 
We compare our relighting capabilities to the single-image relighting approach of~\citet{Zhou_2019_ICCV}.
The source codes of \new{these} techniques are publicly available. 
For Geng~\etal~\shortcite{Geng2018WarpguidedGF}, we estimated the landmarks using the dlib tracker~\cite{dlib09} as suggested by the authors.
We also trained the few shot video-to-video translation method of~\citet{wang2018fewshotvid2vid} for portrait image editing. We trained on 700 videos from the FaceForensics dataset~\cite{roessler2019faceforensics++}. Landmarks were extracted using the dlib tracker as recommended by the authors. 
The approaches of~\citet{Geng2018WarpguidedGF},~\citet{Wiles18} ,~\citet{wang2018fewshotvid2vid} and \citet{Siarohin_2019_NeurIPS} are trained on a video corpus. 
\new{In contrast, our method does not use any direct supervision of the edited images.} 
We compare to these methods in two different settings, self-reenactment and cross-identity reenactment.

\paragraph{Self-Reenactment}
For self-reenactment, we capture several images of a person in different poses. 
We pick the first image and use the other images of the person as reference to edit the head pose.
We captured $9$ people in different poses, resulting in $31$ images in the test set.
Fig.~\ref{fig:pose-comparison} shows some qualitative results. 
\citet{Geng2018WarpguidedGF} use a warp-guided algorithm. While this enables expression changes and in-plane head motion, out-of-plane motion cannot be handled as shown in Fig.~\ref{fig:pose-comparison}.
We also compare to X2Face~\cite{Wiles18}, which samples a learned embedded face in order to synthesize portrait images with different poses and expressions. 
As such, it shares its limitations with~\citet{Geng2018WarpguidedGF} and produces artifacts for strong pose changes. 
\begin{table}[]
\caption{Evaluation of pose edits: We measure landmark alignment errors for same-subject reenactment on 31 images, and facial recognition distances for cross-subject reenactment on 49 images.
Existing landmark detection~\cite{Saragih2009} and facial recognition~\cite{dlib09} often fail on images from competing methods, implying higher realism of our results. }
\begin{tabular}{@{}lll@{}}
\toprule
 & \begin{tabular}[c]{@{}l@{}}Landmark Alignment \\ (number of images)\end{tabular} & \begin{tabular}[c]{@{}l@{}}Recognition \\ (number of images)\end{tabular} \\ \midrule
\citet{Wiles18}    & \textbf{10.9 (22)} & 0.52 (42)          \\
\citet{wang2018fewshotvid2vid}     & 28.19 (24)         & 0.49 (45)          \\
\citet{Siarohin_2019_NeurIPS} & 11.97 (31)         & 0.51 (46)          \\
Ours            & 20.12 (31)         & \textbf{0.40 (49)} \\ \bottomrule
\end{tabular}
\label{tab:pose}
\end{table}
All approaches do not share the same cropping method, which makes it difficult to quantitatively evaluate the results. 
In addition, translation of the head during capture can lead to different illumination conditions. 
Thus, instead of directly computing errors in the image space, we first detect $66$ facial landmarks~\cite{Saragih2009} on all results, as well as the reference images.
We then compute the landmark alignment error, which is the averaged $\ell_2$-distance between the landmarks after 2D Procrustes alignment (including scale). 
The implementation of ~\citet{Geng2018WarpguidedGF} often fails to generate such large pose edits, so we do not consider this approach in the quantitative evaluation. 
Due to artifacts, the landmark detector fails on $29$\% images for the approach of~\citet{Wiles18} and on ~$23$\% for~\citet{wang2018fewshotvid2vid}.
All our results, as well as those of~\citet{Siarohin_2019_NeurIPS} pass through the detector. 
This can be considered as a pseudo-metric of realism, since the landmark detector is trained on real portrait images, implying that our results are better than those of~\citet{Wiles18} and~\citet{wang2018fewshotvid2vid}, and on par with~\citet{Siarohin_2019_NeurIPS}. 
Table~\ref{tab:pose} shows the errors for different methods. 
The low errors for ~\citet{Wiles18} are possibly due to the landmark detector failing in challenging cases.
We obtain only slightly worse results compared to ~\citet{Siarohin_2019_NeurIPS}, even though our method does not have access to ground truth during training. 
~\citet{Siarohin_2019_NeurIPS} train on videos allowing for supervised learning.
In addition, their edits are at a lower resolution of $256 \times 256$, compared to our image resolutions of $1024 \times 1024$.

\paragraph{Cross-identity Reenactment}  We also compare to others in cross-identity reenactment, which is closer to our setting of semantically disentangled editing.  
Here, the image being edited and the reference image have different identities. 
Fig.~\ref{fig:pose-comparison} shows some qualitative results. 
The implementation of~\citet{Geng2018WarpguidedGF} does not support this setting. 
~\citet{Wiles18} and ~\citet{wang2018fewshotvid2vid} result in similar artifacts as discussed before. 
Unlike other approaches,~\citet{Siarohin_2019_NeurIPS} uses two driving images in order to edit the input image, where they use the deformations between the two images as input. 
In the case of self-reenactment, we provide the input image as the first driving image. 
We do the same here, which leads to the two driving images with different identities.
This significantly alters the facial identity in the output image. 
We also quantitatively evaluate the extent of identity preservation for different methods using a facial recognition tool~\cite{dlib09}, see Table.~\ref{tab:pose}.
All methods other than ours do not support semantically disentangled editing. As can be seen in Fig.~\ref{fig:pose-comparison} (bottom), other methods simultaneously change the expressions in addition to the head pose. 
\paragraph{Interactive User Interface} While all existing approaches need a driving image(s) for editing, we allow for explicit editing, using intuitive controls. 
We developed an interactive user interface to edit images, see supplemental video. 
The user can change the head pose using a trackball mouse interface. 
Spherical harmonic coefficients and blendshape coefficients are changed using keyboard controls. 
All editing results run at around 5fps on a TITAN X Pascal GPU.

\paragraph{Relighting}
We compare our relighting results to the single-image relighting approach of~\citet{Zhou_2019_ICCV}, see Fig.~\ref{fig:light-comparison}.
Our approach allows for colored illumination changes, as shown in Fig.~\ref{fig:light-main-ne}. 
Our approach produces higher-quality and more realistic output images. 
We also quantitatively compare the relighting quality of these approaches in an illumination transfer setting, where the illumination in a reference image is transferred to a given input image.  
Since we do not have ground truth data available, we compare the results using a network which predicts the illumination from the reference and the relighted results. 
We use a model-based face autoencoder~\cite{tewari17MoFA}, trained on the VoxCeleb dataset~\cite{Chung18b}.
This network predicts a $27$ dimensional spherical harmonics coefficients. 
We compare the predictions using a scale-invariant $\ell_2$-loss. 
We obtain higher quality ($0.34$), compared to \citet{Zhou_2019_ICCV} ($0.36$).
The numbers are averaged over $100$ relighting results. 
While the method of~\citet{Zhou_2019_ICCV} is only trained for relighting, our method allows us to also edit the head pose and facial expressions.

\new{
\subsection{Generality of the embeddings}
\begin{figure}[t]
\includegraphics[width=0.5\textwidth]{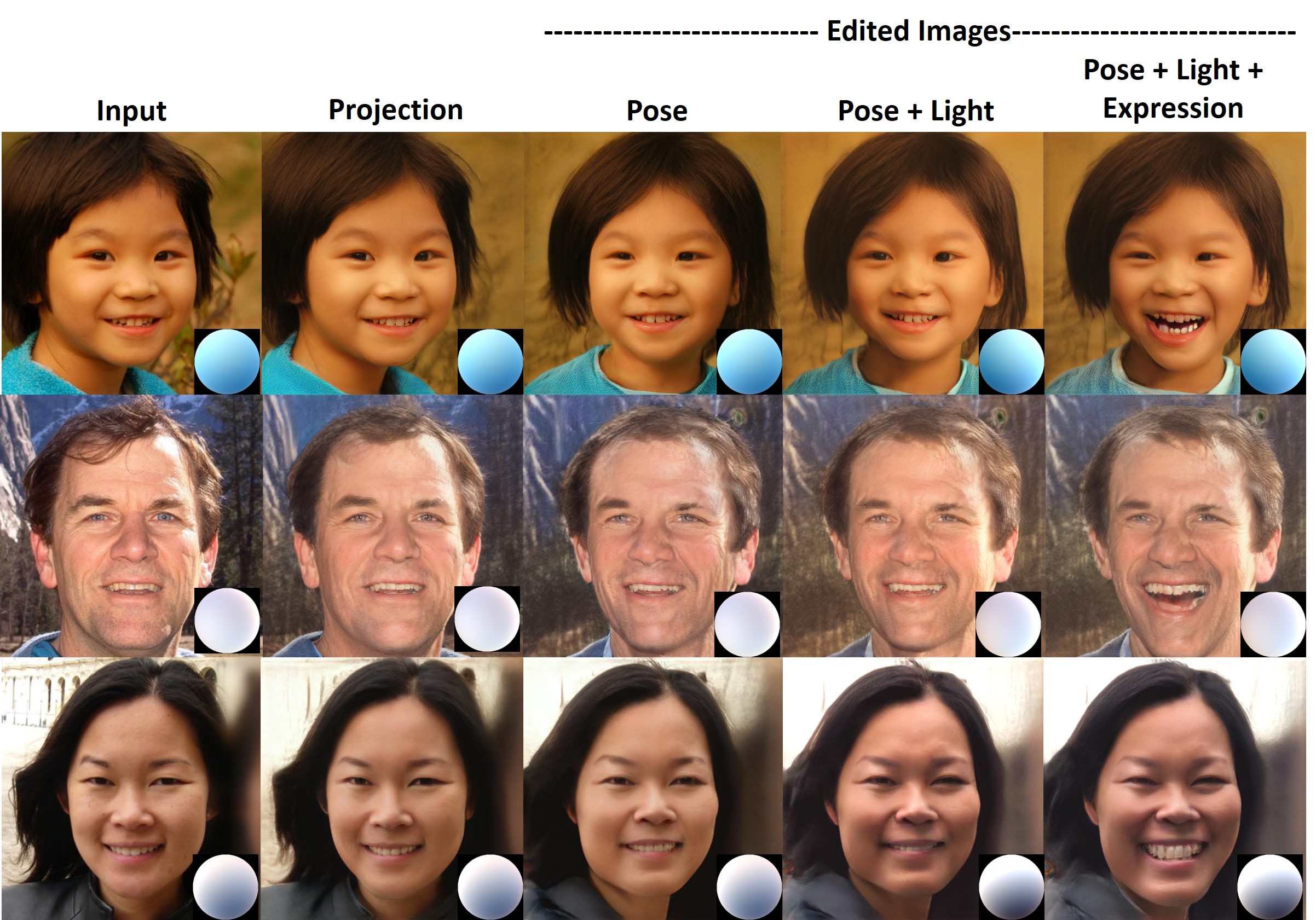}
\vspace{-0.6cm}
\caption{\new{PIE also allows for sequential editing. 
We optimize for the StyleGAN embedding using the pose RigNet. 
We can then use the edited pose results with the RigNets for other semantic components for sequential editing.  Images from~\citet{shen2016deep}.}
}
\label{fig:sequential}
\end{figure}
\begin{figure}[t]
\includegraphics[width=0.5\textwidth]{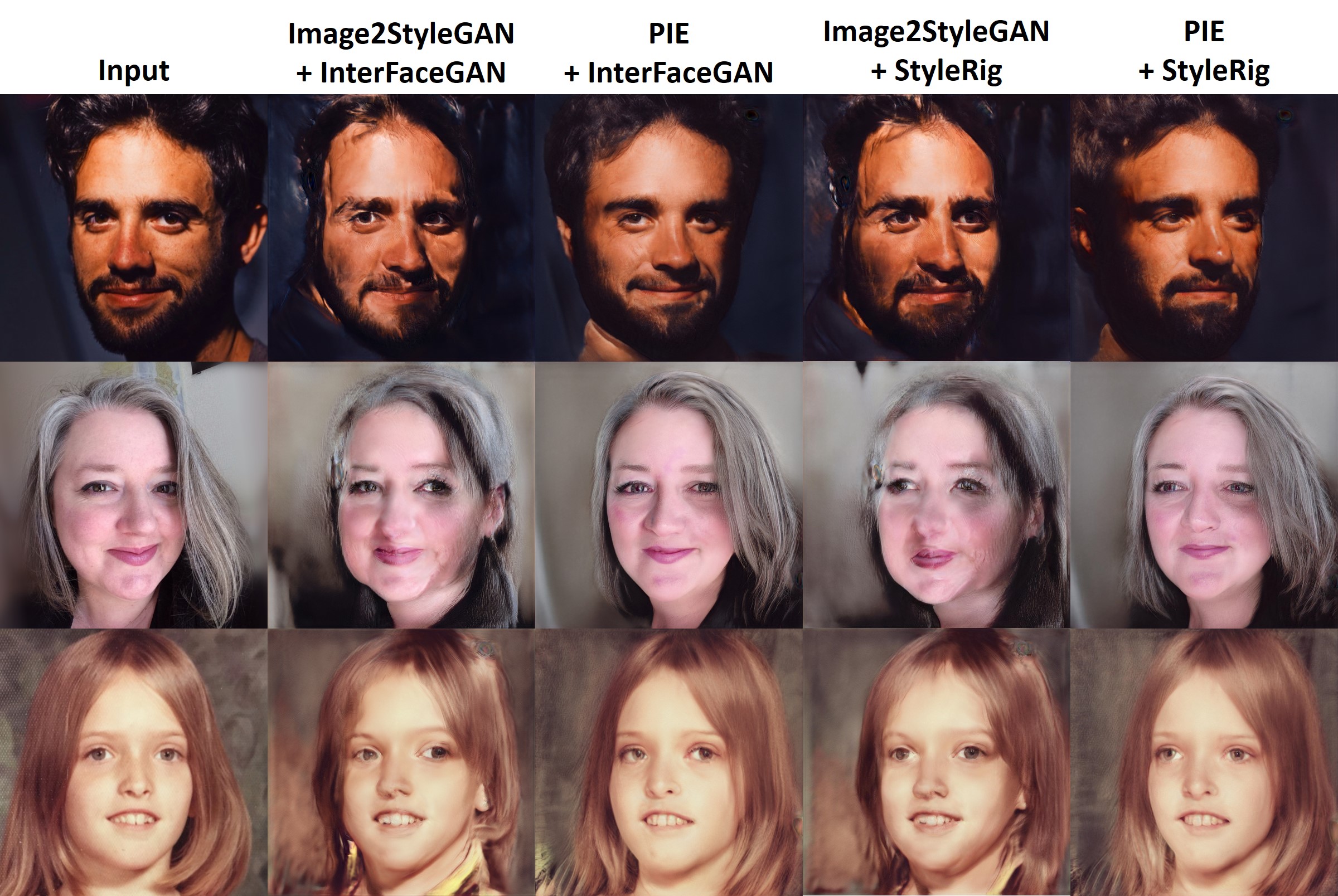}
\vspace{-0.6cm}
\caption{\new{Our embeddings obtain similar quality editing results with the InterFaceGAN~\cite{shen2020interpreting} editing approach. We also notice similar improvements over Image2StyleGAN~
\cite{Abdal_2019_ICCV} embeddings.  Images from~\citet{shen2016deep}.
}}
\label{fig:interfacegan}
\end{figure}
\paragraph{Sequential Editing}
Our method also allows for sequential editing of the different semantic parameters, see Fig.~\ref{fig:sequential}.
Here, we optimize for the embedding using the pose RigNet network. 
After editing the pose, we can use the new embedding as input to the illumination and expression RigNets. 
Since all three versions of RigNet were trained on the same training data, this still produces plausible results.}

\new{
\paragraph{Other StyleGAN editing methods}
Our approach obtains a StyleGAN embedding which can be edited using StyleRig. 
In order to test the generality of these embeddings, we attempt to edit them using InterFaceGAN~\cite{shen2020interpreting}, see Fig.~\ref{fig:interfacegan}. 
Our improvements over Image2StyleGAN generalize to InterFaceGAN editings. 
We better preserve the facial identity and produce fewer artifacts. 
The editing results with InterFaceGAN are of a similar quality to those obtained using StyleRig. 
However, InterFaceGAN cannot change the scene illumination.
}

%% file: 5_limitation.tex
\section{Limitations}
\label{sec:limitation}
\begin{figure}
\includegraphics[width=0.49\textwidth]{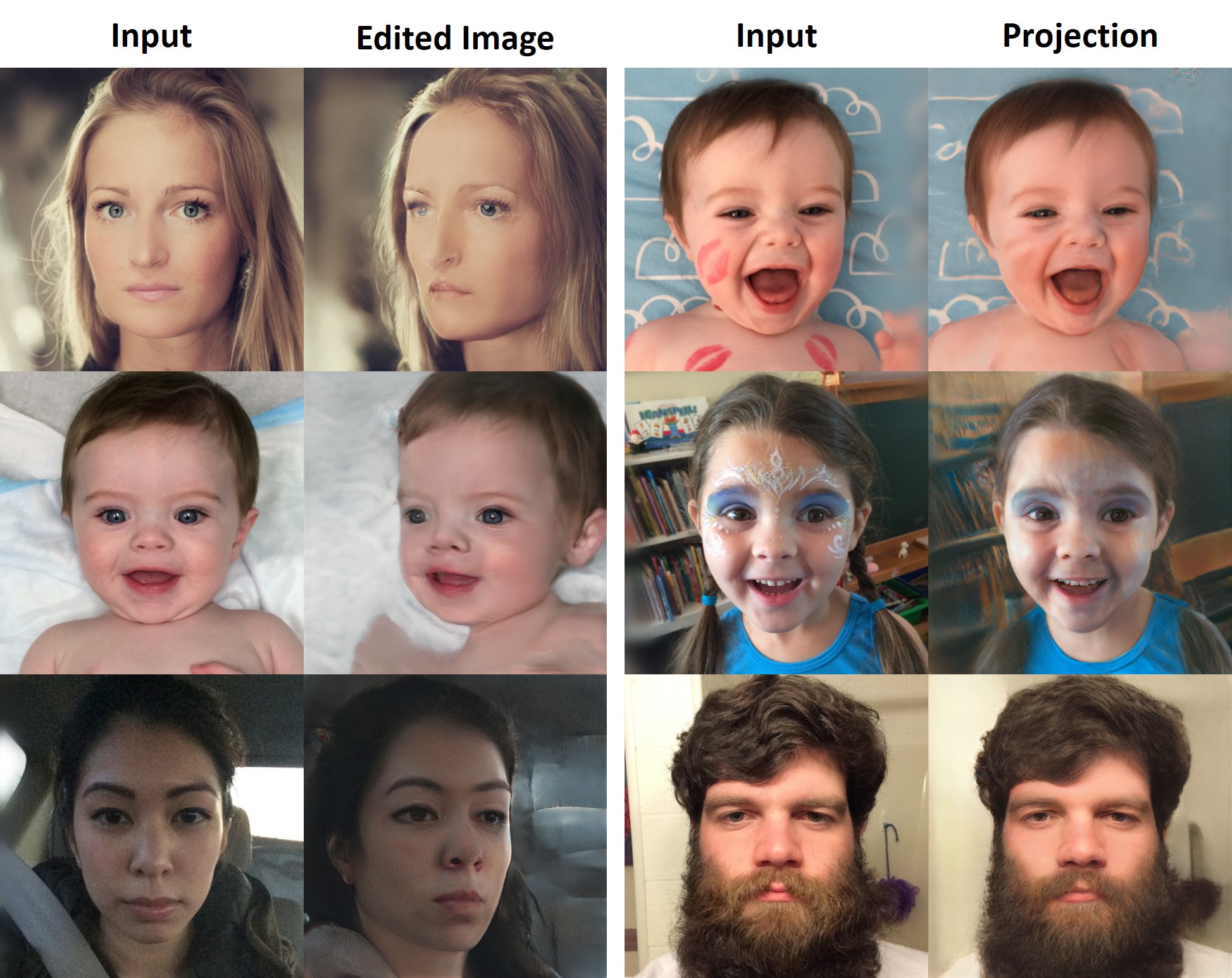}
\vspace{-0.6cm}
\caption{
Limitations: Large edits can lead to artifacts. High-frequency texture on the foreground or background is difficult to fit. Our method also cannot handle cluttered backgrounds or occlusions. Images from~\citet{shen2016deep}.
}
\label{fig:limitations}
\end{figure}
\begin{figure}[t]
\includegraphics[width=0.5\textwidth]{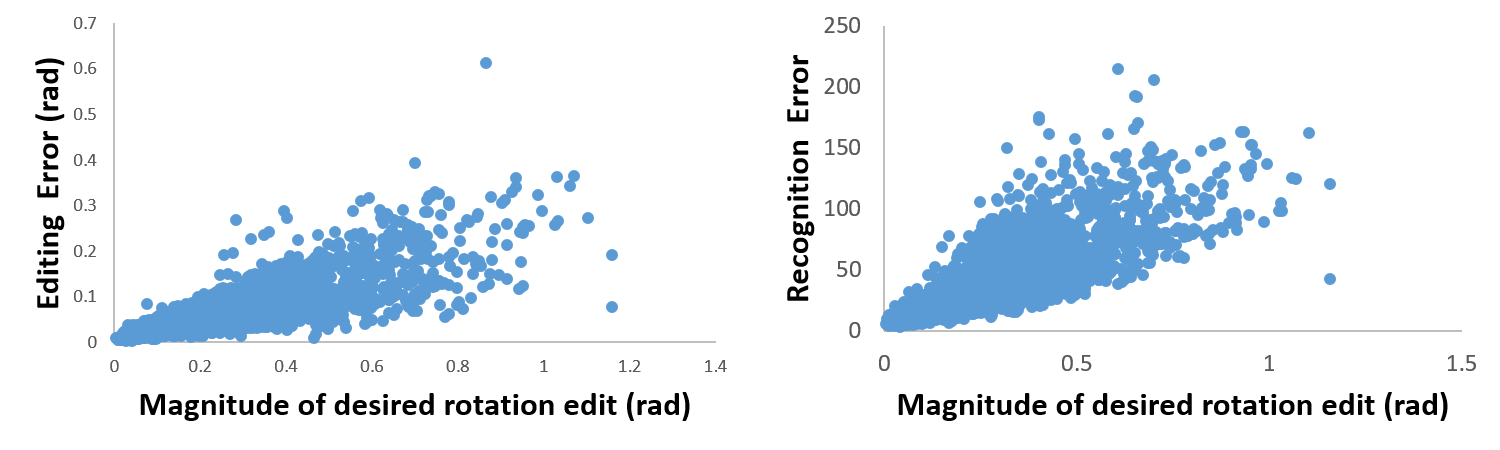}
\vspace{-0.6cm}
\caption{\new{Scatterplot of the editing (left) and recognition errors (right), with respect to the magnitude of the desired pose edits for over $2500$ pose editing results. 
Larger edits lead to both higher editing and recognition errors. 
}
}
\label{fig:large}
\end{figure}

Even though we have demonstrated a large variety of compelling portrait image editing results, there is still room for further improvement of our approach:
(1) At the moment, our approach has a limited expressivity, i.e., it does not allow the artifact-free exploration of the whole parameter space of the underlying 3D morphable face model.
For example, we cannot change the in-plane rotation of the face or arbitrarily change the lighting conditions. 
The main limiting factor is the training corpus (FFHQ \cite{Karras2019cvpr}) that has been used to pretrain the StyleGAN-generator, since it does not contain such variations.
Due to the same reason, our approach is also not yet suitable for video-based facial reenactment, since the variety of facial expressions in the training corpus is severely limited.
This problem could be alleviated by pretraining the generator on a larger and less biased training corpus that covers all dimensions well.
(2) Our method only allows for independent control over the semantic parameters, which is important for editing applications. 
\new{While sequential control is possible, simultaneous control is a more challenging problem.}
(3) Our approach does not provide explicit control over the synthesized background.
At the moment, the background changes during the edits and does not remain static as it should, since the network has learned  correlations between the face and the background.
This could potentially be alleviated by learning an explicit foreground-background segmentation and having a consistency loss on the static background region.
(4) In challenging cases with large deformations, cluttered backgrounds or occlusions and high-frequency textures, our method can fail to faithfully fit to the input image and preserve editing properties at the same time, see Fig.~\ref{fig:limitations}.
In addition, 3D face reconstruction also often fails under occlusions which would lead to incorrect data for our approach.  
\new{
(5) Larger edits generally correspond to worse results, and can often lead to artifacts, as shown in Fig.~\ref{fig:limitations}.
This can also be seen in Fig.~\ref{fig:large}, where larger pose edits correlate with higher editing and facial recognition errors. 
}
(6) Similar to StyleGAN, our approach also sometimes shows droplet-like artifacts.
This could be alleviated by switching to a higher quality generator architecture, such as StyleGAN2 \cite{Karras2019stylegan2}, which has been shown to solve this problem.
\new{
(7) While we show results for people of different ethnicities, genders and ages, we did not extensively study the biases present in the method. 
Some of the components used, such as the 3DMM are known to have racial  biases~\cite{tewari2017self}. 
}
(8) Our results are not guaranteed to be temporally consistent.
While we show temporal editing results (in the supplemental video), our results could be made even more consistent by employing a temporal network architecture and space-time versions of our losses.
Nevertheless, our approach, already now, enables the intuitive editing of portrait images at interactive frame rates.

%% file: 6_discussion.tex
\section{Conclusion}
\new{We have presented the first approach for embedding portrait photos in the latent space of StyleGAN, which allows for intuitive editing of the head pose, facial expression, and scene illumination}.
To this end, we devised a hierarchical optimization scheme that embeds a real portrait image in the latent space of a generative adversarial network, while ensuring the editability of the recovered latent code.
Semantic editing is achieved by mapping the control space of a 3D morphable face model to the latent space of the generator.
In addition, a novel identity preservation loss enables to better preserve the facial identity.

Our approach is a first step towards intuitive and interactive editing of portrait images using a semantic control space akin to computer animation controls.
In addition, our approach provides more insights into the inner workings of GANs, since it allows the intuitive and interactive exploration of the space of face images.
This can shed light  on the biases the model has learned from the employed training corpus.
By using high-quality 3D face models, approaches such as StyleRig would produce better quality with more fine-grained control, and thus would further improve our results. Our paper brings the two different domains of 2D and 3D face models together, thus opening the road towards even more interesting edits. 